%% file: main.tex
\documentclass[preprint,12pt]{elsarticle}

\usepackage{amssymb}
\usepackage{amsmath}

\usepackage{multirow}
\usepackage{url}

\usepackage{blindtext}
\usepackage{titlesec}
\usepackage{graphicx}
\usepackage{subcaption}

\usepackage{algorithm}
\usepackage{algpseudocode}

\usepackage{mathtools}

\makeatletter
\def\ps@pprintTitle{%
 \let\@oddhead\@empty
 \let\@evenhead\@empty
 \let\@oddfoot\@empty
 \let\@evenfoot\@oddfoot
}
\makeatother

\input{commands}

\journal{Artificial Intelligence}

\begin{document}

\begin{frontmatter}



\title{Learning Coupled Subspaces for Multi-Condition Spike Data} 



\author[label1]{Yididiya Y. Nadew}
\author[label2]{Xuhui Fan}
\author[label1]{Christopher J. Quinn}


\affiliation[label1]{organization={Department of Computer Science, Iowa State University},
            city={Ames},
            state={Iowa},
            country={USA}}

\affiliation[label2]{organization={School of Computing, Macquarie University},
                city={Sydney},
                country={Australia}}

\input{abstract}



\begin{keyword}
Latent variable models \sep Spike datasets \sep Gaussian process models \sep Active learning



\end{keyword}

\end{frontmatter}

\input{section__introduction}

\input{section__background_v02}

\input{section__model}

\input{section__inference}
\input{section__prediction_and_active_learning}
\input{section__experiments}

\section{Conclusions}

We proposed \csgpfa, a multi-condition GPFA model that imposes smoothness across latent processes under different experimental conditions. \csgpfa{} 
captures correlations in neural responses and extends GPFA to predict activity under unseen conditions. Leveraging this capability, we explore adaptive experiment design using an active learning algorithm, allowing \csgpfa{} to serve as an exploratory tool for understanding how experimental conditions affect neural responses. Future work includes extensions \csgpfa{} to high-dimensional complex condition spaces (such as natural images, videos, and behavioral inputs). Another direction is incorporating non-stationarity in both latent dynamics and condition dependencies, enabling the model to capture complex dynamics in neural activity.

\newpage
\appendix

\label{sec:appendix:related-works}
\input{appendix__related_works}

\label{sec:appendix:data-augmentation}
\input{appendix__data_augmentation}

\input{appendix__state_space_gaussian_processes}

\input{appendix__variational_updates}

\input{appendix__active_learning}

\input{appendix__experiments}





\bibliographystyle{elsarticle-harv} 
\bibliography{refs}




\end{document}

%% file: commands.tex
\usepackage[english]{babel}
\usepackage[utf8]{inputenc}

\usepackage{colortbl,hhline}
\usepackage{soul}
\usepackage{hyphenat}

\usepackage{graphicx}
\usepackage{amsmath}
\usepackage{amssymb}
\usepackage{amsfonts}
\usepackage{amsthm}
\usepackage{bibentry}

\newcommand{\bs}[1]{\boldsymbol#1} 
\newcommand{\obs}[1]{\overline{\boldsymbol#1}}

\DeclareMathOperator{\Tr}{Tr}

\DeclareMathOperator{\PG}{PG}
\DeclareMathOperator{\diag}{diag}
\DeclareMathOperator{\E}{\mathbb{E}}

\DeclareMathOperator{\Cov}{Cov}

\definecolor{orange}{rgb}{1,0.5,0}

\newcommand{\cnt}[1]{{\bs{#1}_{\textit{c,n,t}}}}

\DeclareMathOperator*{\argmaxA}{arg\,max} %

\newcommand{\blue}{\color{blue}}
\newcommand{\black}{\color{black}}

\usepackage{xspace} %
\newcommand{\csgpfa}{\texttt{CS-GPFA}\xspace}
\newcommand{\nscsgpfa}{\texttt{CS-GPFA(non-smooth)}\xspace}

%% file: abstract.tex
\begin{abstract}

In neuroscience, numerous studies conduct sensory or behavioral experiments under multiple conditions to acquire neural responses in the form of high-dimensional spike train datasets. 
Analyzing high-dimensional spike data %
is a challenging statistical
problem. 
To this end, Gaussian process factor analysis (GPFA), a popular
class of latent variable models, has been proposed for data collected under a single experimental condition. 
GPFA extracts smooth,
low-dimensional latent trajectories that summarize high-dimensional spike datasets. However, standard GPFA infers these trajectories independently for each experimental condition, 
not accounting for how the underlying activity varies across the condition space.
This poses limitations on both accuracy and the interpretability of the latent representation. To address these limitations, we propose Coupled Subspaces GPFA (\csgpfa), 
a Bayesian model that jointly learns latent representations, characterizing how the neural activity varies over the condition space.
Building on this, we further develop an active-learning algorithm for adaptively selecting conditions.
Experiments on both synthetic and real neural datasets demonstrate that \csgpfa achieves superior performance compared to existing approaches. Moreover, our active learning results show that \csgpfa can efficiently guide experiment design in practical settings.

\end{abstract}

%% file: section__introduction.tex
\section{Introduction}
\label{sec:introduction}
In neuroscience, high-density neural probes have enabled scientists to simultaneously record the spiking activities of large populations of neurons \citep{jun2017fully, steinmetz2021neuropixels}. Parallel to the advances in recording techniques, statistical methods have been developed to study the underlying population dynamics \citep{yu2008gaussian, cunningham2014dimensionality, semedo2014extracting, gao2015high, soulat2021probabilistic}. 
A central goal of these methods is to extract simple, low-dimensional representations from noisy, high-dimensional recordings. 
These methods, commonly referred to as latent variable models (LVMs), assume the dynamics of a neural population's activity lies in a low-dimensional latent subspace. 
LVMs provide succinct descriptions of neural activity.

Among diverse classes of LVMs for spike train data, Gaussian process factor analysis (GPFA) has become a widely used extension of classical factor analysis. Originally introduced by \citet{yu2008gaussian}, GPFA combines Gaussian process priors with linear weights to learn smooth trajectories that summarize trial-by-trial spike count data. Subsequent work has extended GPFA to non-Gaussian likelihoods \citep{keeley2020identifying, keeley2020efficient}, to long continuous recordings \citep{jensen2021scalable}, to multiple simultaneously recorded populations \citep{gokcen2023uncovering, li2024multi}, and to efficient Bayesian inference under nonconjugate settings \citep{jensen2021scalable, nadew2024conditionallyconjugate}.

\begin{figure}
    \centering
    
    \begin{subfigure}{0.49\linewidth}
        \centering
        \includegraphics[width=\linewidth]{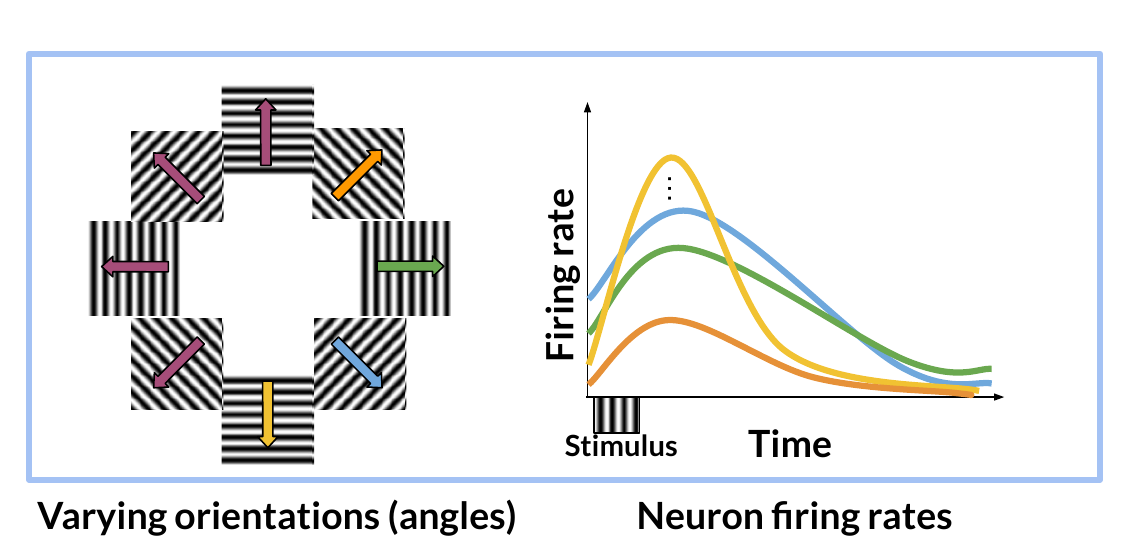}
        \caption{}
    \end{subfigure}
    \hfill
    \begin{subfigure}{0.49\linewidth}
        \centering
        \includegraphics[width=\linewidth]{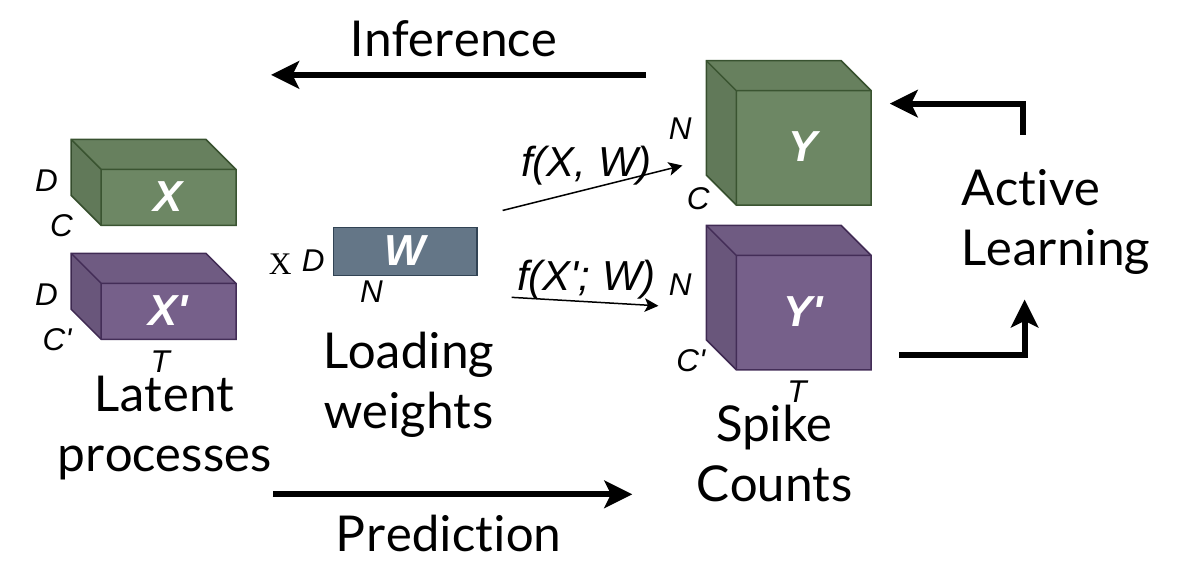}
        \caption{}
    \end{subfigure}

    \caption{
    \textbf{(a)} Conceptual example of drifting gratings experiments with varying orientations (left) and the subject's corresponding neuronal responses (right) in terms of firing rates. 
    \textbf{(b)} \csgpfa schema highlighting three workflows: \textit{inference, prediction, and active learning}. The spike counts, $\bs{Y}$ in green, represent the observations under $C$ conditions. \csgpfa infers the latent trajectories ($\bs{X}$ in green), and predicts of the latent trajectories ($\bs{X}'$) and neural activity ($\bs{Y}'$) under unobserved $C'$ conditions. Based on the predictions, \csgpfa collects new observations via an active learning algorithm. }
    \label{fig:problem}
\end{figure}

While these methods have proven powerful for analyzing activity within a single experimental condition, many neuroscience datasets are collected across multiple, systematically varied conditions. 
For example, numerous studies of visual processing involve sinusoidal static or drifting gratings that vary in contrast, orientation, or speed of motion (see Figure~\ref{fig:problem} (a)) \citep{allen-data, kissinger2018oscillatory, tang2023visual, white2026psychedelic}. 
These conditions form a structured parameter space and often elicit neural responses that are similar but not identical. A major goal when analyzing such datasets is to characterize how changes in experimental conditions produce systematic variations in neural population responses. Another key goal is to predict activity under unseen conditions by interpolating within the condition space, a crucial feature for efficient and adaptive experimental design.

Existing modeling approaches struggle to meet these goals.
GPFA-based models \citep{yu2008gaussian, jensen2021scalable}, despite their flexibility, have typically been used to infer separate representations for each trial (even under a single condition).
Doing so effectively ignores the shared structure among conditions, fails to leverage the parametric nature of the experimental 
conditions, 
and limits interpretability to within-condition dynamics only. 
At the other end of the spectrum, tensor decomposition methods \citep{onken2016using, williams2018unsupervised, soulat2021probabilistic} 
can incorporate multiple conditions by appending the condition index as an additional tensor mode. 
Yet this treatment disregards the parametric geometry of the condition space and lacks the smooth latent trajectories that define GPFA methods.

As a result, neither GPFA nor tensor-based approaches adequately characterize how neural responses evolve over a condition space. %
In addition, post-hoc analysis using existing methods could be problematic in the following regards. 
(1) \textbf{Principled Analysis}---the accuracy of post-hoc analyses using separately fit models may be impaired relative to a jointly fit model.  
(2) \textbf{Sample Complexity}---rich model classes like GPFA often require multiple repeated trials to accurately learn dynamics, often hard or expensive to acquire. 
(3) \textbf{Experimental Design}---due to independent latent structure for each condition, these methods provide limited guidance for designing experiments under varying conditions.
To address these challenges, we propose a novel GPFA model called 
\textit{coupled subspaces GPFA}, or \texttt{CS-GPFA} for short. 
\texttt{CS-GPFA} analyzes data from multiple experimental conditions jointly (hence ``coupled'') by exploiting the parametric nature of the conditions. \csgpfa unifies three main workflows: \textit{inference} of latent structure for multi-condition data, \textit{prediction} of neural activity under unobserved conditions, and \textit{adaptive selection} of experimental conditions (see Figure~\ref{fig:problem} (b)).  
We summarize our contributions as follows, 

\begin{itemize}
    \item We propose a novel GPFA model
    to jointly analyze neural datasets under multiple experimental conditions. Here, we show how we can exploit the label information on the conditions to learn coupled latent representations of neural datasets.  
    
    \item We develop a scalable Bayesian inference procedure via a data augmentation strategy to derive simple, closed-form updates for our model posteriors. Leveraging the state-space dual representation of GPs, we derive an efficient, linear-time inference procedure for long recordings.

    \item We develop an active learning algorithm to adaptively select conditions.

\end{itemize}

\noindent
We empirically demonstrate the utility of our method on both synthetic and real-world datasets
under varying low-data settings, such as few trials per experimental condition and few experimental conditions.

We begin in Section~\ref{sec:related-works} with a discussion of the relevant previous works and highlight their key similarities/differences with our method. In Section~\ref{sec:model}, we present our novel \csgpfa model with assumptions and Bayesian priors. We develop an efficient inference algorithm in Section~\ref{sec:inference} and adaptive experimental design in Section~\ref{sec:prediction_and_active_learning}, and Section~\ref{sec:experiments} presents our empirical results on real-world neural datasets.

%% file: section__background_v02.tex
\section{Related Work}
\label{sec:related-works}
In this section, we briefly review relevant literature on popular latent variable models (LVMs) such as GPFA models and tensor decomposition methods.  
See \ref{apndx:related-works-extended} 
for more discussion.

\textbf{GPFA Models}
First proposed by \citet{yu2008gaussian}, the GPFA model unifies the smoothness of Gaussian processes with dimensionality reduction into a probabilistic framework.
To the best of our knowledge, no prior work has proposed a GPFA model to capture variations across experimental conditions, though several works consider related problems.
\citet{yu2008gaussian}'s model fits latents separately for each trial (exploiting shared weights and hyperparameters), which provides no (explicit) coupling of the latents across conditions.
\citet{keeley2020identifying} accounted for trial-to-trial variability in neural datasets by splitting the latent structure into ``signal'' (shared among trials) and ``noise'' (unique to trials). They did not explicitly consider multiple conditions, and directly applying their approach to multiple conditions data could misattribute condition-driven changes to trial noise. %
In our work, we consider trials under a single condition as i.i.d.~with smoothly varying latents over the condition space.  
Several works have proposed GPFA models for multi-population recordings by identifying lead-lag relationships between the populations \citep{gokcen2023uncovering, li2024multi, gokcen2025fast, li2025learning}. That direction is complementary to ours, as we focus on a single population recorded under varying conditions, where the variability extends beyond simple temporal shifts. %

In terms of model design and inference, the closest related work is  \citet{nadew2024conditionallyconjugate}.  They considered a single population recorded under a single condition.  They developed a conditionally conjugate GPFA (ccGPFA) which leverages a data augmentation technique to render the model conditionally conjugate. 
This results in closed-form solutions for the conditional posteriors in the negative-binomial likelihood case.  In this work, we consider a more challenging setting, multi-condition modeling, but show data augmentation can yield conditional conjugacy, enabling closed-form updates. Unlike \citet{nadew2024conditionallyconjugate}, whose latent functions are one-dimensional GPs, we model multi-dimensional latent functions, making a direct generalization computationally prohibitive, with complexity scaling as $\mathcal{O}((CT)^3)$ for $C$ conditions and $T$ time points. In this work, we develop an efficient update rule that scales linearly w.r.t. time points. 
In addition, via joint modeling of multiple conditions, our method naturally supports tasks such as prediction under unobserved conditions and active learning.

%

%

%
%
%
%

%
%
%

%
%
%
%
%
    
    %
    %
%

%

%

%
%
%
%
%
%
%
%
%
%
%
%
%
%
%
%
%
%
%
%
%
%
%
%
%
%
%
%
%
%
%
%
%
%
%
%
%
%
%
%
%
%
%
%
%
%
%
%
%
%
%
%
%

%

%
%
%
%
%

%
%
%

%

\textbf{Tensor-based Methods for Neural Data}
Another line of work for neural data analysis is based on a family of tensor decomposition (TD) models.
These methods allow analyzing data with a set of dimensions, such as the number of neurons, recording bins, and trials. \citet{onken2016using,williams2018unsupervised, soulat2021probabilistic} fold multi-trial neural datasets into a tensor and apply tensor decomposition. 
A major contrast between tensor-based methods and GPFA based methods is that tensor-based methods do not enforce 
smoothness along any dimension, in particular along time or condition.
In this work, we preserve the smoothness property and consider experimental conditions explicitly in our model. Consequently, our method not only yields coupled latent representations but also predictions under unobserved experimental conditions and adaptive experiment design.

%% file: section__model.tex
\section{\csgpfa Model}
\label{sec:model}

Here, we introduce our proposed model, \textit{coupled subspaces GPFA} (\texttt{CS-GPFA}). We formulate a Bayesian model, discuss likelihood choices, and outline the specialized latent structure for multi-condition spike data. %

Consider a tensor of spike trains $\bs{Y} \in \mathbb{N}^{C \times N \times T}$ of a set of $N$ simultaneously recorded neurons under $C$ different experimental conditions. %
Each spike train is assumed to have been (synchronously) binned into $T$ uniformly-sized bins (a.k.a. `time steps'). %
Our goal is to learn a low-dimensional latent structure that summarizes the neural activity across both \textit{time} and \textit{conditions}.

We model the spike trains $\bs{Y}$ as arising from a linear mixing of low-dimensional $D$ latent structures $\bs{X} \in \mathbb{R}^{C \times D \times T}$, where $D \ll N$. The entry $\bs{X}_{c,d,t}$ represents the $d$-th latent activity for condition $c$ at time step $t$.
The neural activity is linked to these latents via a loading matrix 
$\bs{W} \in \mathbb{R}^{N \times D}$, where $\bs{W}_{n,d}$ represents the contribution of the latent dimension $d$ to neuron $n$. Furthermore, we consider a bias term for each neuron $\bs{\beta} \in \mathbb{R}^{N \times 1}$ to capture its base activity.

\black 

\subsection{Likelihood}

Extending the conditional independence assumption for likelihood in \citet{yu2008gaussian}, the likelihood of the tensor spike count data $\bs{Y}$ given all latent variables factorizes across all dimensions (conditions, neurons, and time steps). As discussed in Section~\ref{sec:related-works},
the choice of the likelihood distribution is crucial in modeling neural datasets. 
To account for under-dispersion \citep{maimon2009beyond, kara2000low}  and over-dispersion \citep{pillow2012fully, gao2015high} in the variance of spike counts common in neural data, 
we use binomial and negative binomial likelihoods for our model, respectively.

For the negative binomial likelihood model, we consider distinct dispersion parameters (capturing the ``number of failures'')
$\{ \bs{r}_n \}_{n=1}^N$ shared across experimental conditions and time. 
With the latent combination $\cnt{F} = \bs{\beta} + \sum_{d=1}^{D} \bs{W}_{n,d} \bs{X}_{c,d,t} $, the likelihood of the spike data $\bs{Y}$ %
factorizes as %

\begin{align} 
    &%
    p(\bs{Y} | \bs{W}, \bs{X}, \bs{\beta_n}, \{ \bs{r}_n \}) = \prod_{c,n,t} \frac{  \Gamma(\cnt{Y} + \bs{r_n})}{\cnt{Y}!\Gamma(\bs{r_n}) } \frac{\exp(\cnt{F})^{\cnt{Y}}}{  ( 1 + \exp(\cnt{F}))^{{\cnt{Y}} + \bs{r}_n} } , \label{eq:negative-binomial-likelhood}
\end{align} 
where $\Gamma$ is the gamma function.\footnote{We use $\prod_{c,n,t}$ as a shorthand for  $\prod_{c=1}^C \prod_{n=1}^N \prod_{t=1}^T $.} While we work with a negative binomial likelihood model, with minimal changes, our method also applies to a binomial likelihood model (see \ref{sec:appendix:augmentation:binomial}).

\subsection{Latents as functions over conditions and times}

    \label{sec:likelihood:latents}

    As noted in Section~\ref{sec:introduction}, previous GPFA methods \citep{yu2008gaussian, keeley2020efficient,jensen2021scalable}
    do not explicitly account for experimental conditions. 
    One approach to apply them would be to
    analyze data from multiple conditions separately (i.e., for each condition, they consider a separate pool of latent processes).
    With that, the posterior distributions would become effectively independent across conditions. 
    In this work, assuming that each condition is parameterized with a set of continuous variables, 
    we define the latent processes as \textit{smooth continuous functions} over the condition space. The key idea is to induce a correlation among latent processes across experimental conditions. To model these processes, we choose multi-input Gaussian process (GP) priors by augmenting the temporal space with space of experimental conditions.

    In particular, we model the $d$-th  latent process, $\bs{X}_{:,d,:} \in \mathbb{R}^{C \times T}$, under $C$
    conditions as drawn from a multi-input single-output GP distribution \citep{rasmussen2006gaussian}.
    Following standard practice, we place a non-informative zero mean function for our prior. We let $ \kappa (t, t', \bs{c}, \bs{c}')$ denote the kernel function that evaluates the covariance between latent variables at {points $\{ t, \bs{c}\}$  and $\{ t', \bs{c'}\}$ where $t, t'$ are timestep indices and $\bs{c'}$ and $\bs{c}$ represent two points in the condition space and time. Then
    \begin{align}
       \bs{X}_{:,d,:} \sim \mathcal{GP} ( \bs{0}, \kappa (t, t', \bs{c}, \bs{c}')). \label{eq:prior:latent-gps}
    \end{align}
    } 
    
    We choose a separable product kernel function along temporal and condition dimensions, i.e.  $\kappa (t, t', \bs{c}, \bs{c}') = \kappa(t, t') \kappa(\bs{c}, \bs{c}')$.  
    For each kernel function, we choose a kernel from the Mat\'{e}rn class of covariance functions.
    This is motivated by their equivalent state-space formulation \citep{hartikainen2010kalman,sarkka2013spatiotemporal}, which lays the foundation for efficient inference. We will shortly show how this is useful for both inference and prediction. 

For concise modeling, we integrate the bias terms $\bs{\beta}$ as an additional latent dimension with a fixed constant mean of $\bs{1}$ and zero variance.

    \subsection{Priors over remaining variables}
    \label{subsection:method:more-priors}
    Similar to \citet{nadew2024conditionallyconjugate}, 
    we put an independent multivariate Gaussian prior on the weights: $p(\bs{W}) = \prod_n \mathcal{N}(\bs{0}, {\diag(\bs{\tau}^{-1})})$ where $\bs{\tau} \in \mathbb{R}_+^{D \times 1}$ denotes the precision parameters corresponding to $D$ the latent dimensions.
        
    We place an \textit{automatic relevance determination} (ARD) prior on the precision parameters, $p(\bs{\tau}) = \prod_d \Gamma(\alpha_d, \beta_d)$ where $\alpha_d$ and $\beta_d$ are the shape and rate parameters of the gamma distribution. This prior penalizes the total number of non-zero columns, encouraging the sparsity in the columns of the weight matrix.

    For the dispersion variables ($\bs{r}_n$), we treat them as model parameters and initialize them using the mean spike counts of the respective ($n$-th) neurons in the dataset. 

\black

%% file: section__inference.tex
\section{Inference}
\label{sec:inference}

The negative binomial likelihood in \eqref{eq:negative-binomial-likelhood} results in a non-conjugate Bayesian model that is analytically intractable. In addition, computing the posterior using off-the-shelf sampling approaches, such as Markov Chain Monte Carlo (MCMC), can be prohibitively slow due to the large number of variables in the model. To address this challenge, we introduce auxiliary variables that restore conditional conjugacy, facilitating inference. Based on this augmentation, we propose an efficient inference algorithm.

\subsection{Restoring conjugacy via data augmentation}

We first augment auxiliary variables to the \texttt{CS-GPFA} model likelihood to render it conditionally conjugate. 
To do so, we transform each exponential term in the likelihood \eqref{eq:negative-binomial-likelhood} 
using integral identities 
presented in 
\citet{polson2013bayesian},
    \begin{align}
        \frac{\exp(\cnt{F})^{\cnt{Y}}}{  ( 1 + \exp(\cnt{F}))^{{\cnt{Y}} + \bs{r}_n} } &= 2^{-(\cnt{Y} + \bs{r}_n )} \nonumber \\ &\quad \times \int_0^\infty \!\!\!\!\exp( \frac{(\cnt{Y}+ \bs{r}_n )\cnt{F} }{2} -\frac{\cnt{\omega} \cnt{F}^2}{2})  \nonumber  \\
        & \quad \times \text{PG} (\cnt{\omega}; \cnt{Y}+ \bs{r}_n, 0) d \cnt{\omega}. 
    \end{align}
    where $\text{PG}(\cnt{\omega}; \cdot,\cdot)$ refers to the Pólya-gamma distribution for the auxiliary variable $\cnt{\omega}$.
    
    Conditioning on the newly introduced auxiliary variables $\{\bs{\omega}_{c,n,t}\}$ into our model for all $ c \in [C]$, $n \in [N]$ and $ t \in [T]$, 
    the original negative-binomial likelihood over $\bs{Y}_{c,n,t}$ reduces to an unnormalized Gaussian likelihood over a transformed variable  $\hat{\bs{Y}}_{c,n,t} = \frac{\bs{Y}_{c,n,t} - \bs{r}_n}{2 \bs{\omega}_{c,n,t}}$. 
\black
In addition, the full conditional distributions of model variables $\{\bs{X}, \bs{W}, \bs{\tau} \}$   and all auxiliary variables $\{\bs{\omega}_{c,n,t}\}$ are available in closed form (see \ref{sec:appendix:augmentation:augmentation} for details).

\subsection{Efficient free-form variational inference}
Given the availability of full conditionals, the augmented model naturally lends itself to Gibbs sampling \cite{casella1992explaining}. 
Although it is easier to implement, this is impractical due to its slow convergence. Instead, we propose a fast variational EM algorithm to learn the posterior distribution of our model variables. 

     Using the mean-field assumption, we approximate the joint posterior distribution over
     $\bs{\Theta} = \{ \{ \bs{\omega}_{c,n,t} \}, \bs{X}, \bs{W},\bs{\tau}  \} $, 
     compactly denoted as $p(\bs{\Theta} | \bs{Y})$,
     by choosing a restricted and factorized, yet tractable, variational distribution $q(\bs{\Theta})$ as:
        \begin{align}
            q(\bs{\Theta}) = q(\bs{W})  q(\bs{\tau}) \prod_{c,n,t} q(\bs{\omega}_{c,n,t}) \prod_{d} q(\bs{X}_{:, d, :}).\label{eq:variational-family-dist}
        \end{align}
We make no further assumptions on the functional form of the $q(\bs{\Theta})$ distribution. %
We optimize $q(\bs{\Theta})$ with respect to the lower bound of the marginal log-likelihood $\log p(\bs{Y})$ called the evidence lower bound (ELBO), which is denoted $\mathcal{L}(q)$ as
\begin{align}
        \mathcal{L}(q) := \E[\log p(\bs{Y} | \bs{\Theta}) p(\bs{\Theta}) /q(\bs{\Theta})].
\end{align}
\textbf{E-step} In this step, we iteratively solve the above objective by fixing all $q$ distributions in Eq.~(\ref{eq:variational-family-dist}) except one and optimizing the objective w.r.t. it. From \citep{bishop2006pattern} (Chapter 10.1),  we know the optimal variational distribution for a variable $\Theta_j \in \bs{\Theta}$, denoted as $q^*(\Theta_j)$, is given by  
\begin{align}
    q^*(\Theta_j) \propto \exp\left\{ \E_{q(-\Theta_j)} [\log p(\bs{Y}| \bs{\Theta}) p(\bs{\Theta})]\right \} .
    \label{eq:var-update-equation}
\end{align}
\black
Here the expectation is taken w.r.t. the product of all of the variational distributions except $q(\Theta_j)$. 
Using \eqref{eq:var-update-equation}
for each $q$ distribution in \eqref{eq:variational-family-dist}, 
we derive its optimal variational distribution. 

\label{subsec:inference:optimal-latents}
\noindent \textbf{Optimal latent processes $q^*(\bs{X}_{:, d, :})$} 
   A naive implementation of the derivation results in $O((CT)^3)$ complexity with $C$ conditions and $T$ timesteps, mainly due to the cost of inverting $CT \times CT$ covariance matrix. 
   To address this issue, we leverage the separability of the time and condition kernels and state-space representation of GP kernels to develop an efficient update rule. 
   First, we construct an equivalent state-space formulation by augmenting the
   time derivatives
   $\{\frac{\partial\bs{X}_{:,d,:}}{\partial{t}}\}_d$  
        
   into the model \citep{hamelijnck2021spatio} 
   (see details in \ref{sec:appendix:state-space-gp-formulation}). The prior distribution can be written as $p(\bs{X}_{:,d,:}) = \int p(\bs{X}_{:,d,:}, [\frac{\partial\bs{X}_{:,d,:}}{\partial{t}}]) d [\frac{\partial\bs{X}_{:,d,:}}{\partial{t}}]$. 
   Denoting the state at time $t$, $\overline{\bs{X}}_{:,d, t} \in \mathbb{R}^{C\hat{d}}$, where $\hat{d}$ is the dimensionality of the state induced by $\kappa(t, t')$,  the state-space model is given as
   \begin{align}
       \overline{\bs{X}}_{:,d, t+1} &= \left[ \bs{I}_C \otimes  \bs{A}_t \right] \overline{\bs{X}}_{:,d, t} + \bs{q}_t, \tag{state transition}  \\ %
       \bs{X}_{:,d, t+1} &= \left[ \bs{I}_C \otimes \bs{H} \right] \overline{\bs{X}}_{:,d, t+1}, \tag*{measurement}
       \label{eq:state-space-prior-formuation}
   \end{align}
where $\bs{q}_t \sim \mathcal{N}(0, K_{CC}^{(\bs{c})} \otimes \bs{Q}_t)$. 
Here, $\bs{A}_t$, $\bs{Q_t}$, $\bs{H}$ are the transition matrix, process noise covariance, 
and measurement model induced by the temporal kernel $\kappa(t, t')$ (see \ref{sec:appendix:state-space-gp-formulation} for details on the matrices for Mat\'{e}rn kernels). 
$K_{CC}^{(\bs{c})}$ denotes the covariance matrix induced by the condition kernel $\kappa(\bs{c}, \bs{c}')$ on $C$ condition points.
Note that with this Markovian structure, for $t'<t-1$, $\overline{\bs{X}}_{:,d, t}$ is conditionally independent of $\overline{\bs{X}}_{:,d, t'}$ given $\overline{\bs{X}}_{:,d, t-1}$. We rewrite  \eqref{eq:prior:latent-gps}
    
as $p(\overline{\bs{X}}_{:,d, :}) = p(\overline{\bs{X}}_{:,d, 1}) \prod_{t=1}^T p(\overline{\bs{X}}_{:,d, t} | \overline{\bs{X}}_{:,d, t-1})$. Hereafter, we denote the set of augmented latent processes, $\{ \overline{\bs{X}}_{:,d, :} \}$ simply as $\overline{\bs{X}}$.

Specifically solving for $q^*(\bs{X}_{:, d,t})$ using the mean-field equation, we recognize it as an unnormalized Gaussian \textit{pseudo likelihood} multiplied by the above formulated state-space prior.
    
We avoid computing its full posterior distribution.  
Instead we compute the marginal posteriors $\{q^*(\overline{\bs{X}}_{:, d,t}) = \mathcal{N}(\bs{m}_{ d,t}, \bs{P}_{d,t})\}_{d,t}$ using  Kalman filtering
\citep{kalman1960new} and Rauch–Tung–Striebel smoothing \citep{rauch1965maximum}. 
In addition, using the deterministic relationship between $\bs{X}_{:,d, t+1}$ and its corresponding state \eqref{eq:state-space-prior-formuation}, 
the marginal posterior is %
\begin{align*}
q^*(\bs{X}_{:,d, t + 1}) = \mathcal{N}( \left[ \bs{I}_C \otimes \bs{H} \right] \bs{m}_{d, t+1}, \left[ \bs{I}_C \otimes \bs{H} \right] \bs{P}_{d, t+1} \left[ \bs{I}_C \otimes \bs{H} \right]^\top ).
\end{align*}
For the remaining variational distributions, careful derivations reveal that the marginal posteriors $\{ q^*(\bs{X}_{:, d,t})\}$ are sufficient for their updates. 
Therefore, it is not necessary to compute its full posterior covariance.

Using this state-space formulation, we significantly reduce the computation complexity from $O((CT)^3)$ to only be linear w.r.t. time steps $T$ and only cubic w.r.t. the number of conditions $C$ and the state dimension: $O(T (\hat{d}C)^3)$. While similar state-space formulations have been applied to spatio-temporal problems \citep{hamelijnck2021spatio, zhu2023markovian}, their use with non-conjugate count likelihood here is, to our knowledge, particularly novel. This is important as the previous methods handled non-conjugate likelihood via projection to a conjugate family \citep{hamelijnck2021spatio}, or Gaussian site approximation \cite{zhu2023markovian}. In contrast, we use augmentation to sidestep non-conjugacy, without approximating the likelihood.
Due to limited space, we defer the complete details of the derivations and update rules for all variables to \ref{sec:appendix:variational-updates}.  %

\textbf{M-step } After cycling through the model variables in $\bs{\Theta}$ in the E-step, we optimize our model hyperparameters, such as the GP parameters,  using the Adam optimizer module from the PyTorch library \citep{paszke2019pytorch}. In addition, we optimize the dispersion parameters, $\{ \bs{r}_n \}$, with respect to the ELBO objective. 

%% file: section__prediction_and_active_learning.tex
\section{Prediction and Active Learning}
\label{sec:prediction_and_active_learning}

\input{section__prediction}

\input{section__active_learning}

%% file: section__prediction.tex
\subsection{Predicting neural activity under novel conditions}

    In practice, researchers often conduct experiments over a set of finite conditions. 
    For some experiments, an important question could be \textit{given a set of observed conditions and their corresponding neural responses, is it possible to predict the responses for unseen conditions?} 
    For our model, this problem reduces to predicting the posterior distribution of
    a new set of latent processes for the unseen conditions. 
    This follows from the fact that the weights are shared across conditions and their corresponding latent processes are defined as smooth functions of the condition labels (locations in the condition space). 
    Our GP assumption over the latent functions
    \eqref{eq:prior:latent-gps} allows us to predict the latent states for new conditions and model their uncertainty.

    To do so, we make a common assumption that $\bs{\Theta}$ is a set of sufficient statistics for the new set of latent processes, i.e., the posterior over $\bs{\Theta}$ captures the posterior over the new latent processes. 
    Denoting the latent processes $\bs{X}'$ over a set of $C'$ (unseen) conditions, the joint posterior distribution $p(\bs{X}', \bs{\Theta} | \bs{Y})$ can be approximated as $q(\bs{X}', \bs{\Theta}) := p(\bs{X}'| \bs{\Theta}) q(\bs{\Theta})$. Using the prior independence assumption \eqref{eq:prior:latent-gps}, 
    $\bs{X}' \perp  (\bs{\Theta} \setminus \bs{X}) | \bs{X}$,

    and posterior independence in \eqref{eq:variational-family-dist}, the marginal posterior 
    \[q(\bs{X}') = \prod_d \int p(\bs{X'}_{:,d,:}| \bs{X}_{:,d,:}) q(\bs{X}_{:,d,:}) d\bs{X}_{:,d,:} \ .\]
    Letting  $q(\bs{X}_{:,d,:}) = \mathcal{N}(\bs{m}, \bs{P})$ with posterior mean $\bs{m}$ and covariance $\bs{P}$, the posterior mean $\bs{m}'$ and covariance  $\bs{P}'$ for the new points $\bs{X}'$ are given by,

\begin{align}
    &\bs{m}' = 
    \left[
      I_T \otimes 
      \left(
        K^{(\bs{c})}_{C'C}\,(K^{(\bs{c})}_{CC})^{-1}
      \right)
    \right]
    \bs{m}
    \tag{predicted mean} \\
    &\bs{P}' = 
    \left[
      K^{(t)}_{tt} \otimes
      \left(
        K^{(\bs{c})}_{C'C'}
        - K^{(\bs{c})}_{C'C}\,(K^{(\bs{c})}_{CC})^{-1}\,K^{(\bs{c})}_{C C'}
      \right)
    \right] \nonumber\\
    &+ \left[
      I_T \otimes 
      \left(
        K^{(\bs{c})}_{C'C}\,(K^{(\bs{c})}_{CC})^{-1}
      \right)
    \right]
    \bs{P}
    \left[
      I_T \otimes
      \left(
        (K^{(\bs{c})}_{CC})^{-1}\,K^{(\bs{c})}_{C C'}
      \right)
    \right]. 
    \tag{predicted covariance}
\end{align}

where matrices with superscripts $(\bs{c})$, $(t)$ denote the covariances between condition points and timesteps respectively. Recall that we do not know the full form of $q(\bs{X}_{:,d,:})$, 
but only its marginal distribution with respect to the time axis (i.e $P$ is only partially known ) (see Section~\ref{sec:inference}). However, due to the block diagonal structure of $I_T \otimes
  (
    K^{(\bs{c})}_{C'C}\,(K^{(\bs{c})}_{CC})^{-1}
  )$, as shown in \citep{hamelijnck2021spatio}, we can isolate the marginal distributions for the predicted latents $q(\bs{X'}_{c,d,t})$. %
    Using Lemmas I.1 and I.2 in
    \citep{hamelijnck2021spatio}, the marginal $q(\bs{X'}_{c,d,t})$ is a Gaussian:
    $q(\bs{X'}_{c,d,t}) = \mathcal{N}\big(\bs{m'}_{c,t}, \bs{P'}_{c,t}\big),$
    where the mean and covariance are
    \begin{align}
    \bs{m'}_{c,t} &= (K^{(\bs{c})}_{cC} (K^{(\bs{c})}_{C C})^{-1})  \bs{m}_t, \\
    \bs{P'}_{c,t} &= \, K^{(\bs{c})}_{c c}K^{(t)}_{tt}
               + K^{(\bs{c})}_{c C} (K^{(\bs{c})}_{C C})^{-1} \left [- K^{(t)}_{tt} K^{(\bs{c})}_{C c} 
               + \bs{P}_t \, K^{(\bs{c})}_{C C} \, K^{(\bs{c})}_{Cc} \right ]. \label{eq:latent-prediction}
    \end{align}
Here, $\bs{P}_t$ is the $t$-th block in the posterior distribution, which we recover from the inferred marginal posterior at time $t$ via Kalman filtering and smoothing.  

With the predicted latent trajectories $q(\bs{X'}_{c,d,t})$ under novel conditions and the shared loading weights, the neural activity under $c$, or equivalently the posterior predictive distribution $p(\bs{Y}'| \bs{Y})$ can be computed via simple Monte-Carlo approximation (See \ref{subsec:appendix:posterior_prediction} for complete details).

%% file: section__active_learning.tex
\subsection{Active learning: Efficient exploration of the condition space }
\label{sec:active-learning}

Building on the capability to predict,
here we develop a technique for adaptive experimental design. %
In particular, we attempt to answer the question: \textit{given a space of conditions and limited budget, how can one adaptively
select experimental conditions %
that sufficiently explore the space while satisfying the budget constraint?} We define the budget as the number of distinct conditions to use in the experiment. To answer this question, we develop an active learning algorithm based on two objectives: uncertainty sampling (US) \citep{lewis1994sequential, yang2016active}, and information gain (IG) \citep{houlsby2011bayesian}.

\begin{algorithm}[t]
    \label{alg:active-learning-1}
    \caption{Active Learning using \texttt{CS-GPFA} }
    \begin{algorithmic}
        \Require Set of all conditions $\mathcal{C}$, data under a subset of training conditions $\mathcal{C}_{train} \subset \mathcal{C} $, %
        budget $\mathcal{B}$
        \While { $|\mathcal{C}_{train}| < \mathcal{B} $ }  
        
            \State Fit a \texttt{CS-GPFA} model $M_{\mathcal{C}_{train}}$ using all observations $(\{\bs{Y}_{c,:,:}\}_{c\in \mathcal{C}_{train} })$ under $\mathcal{C}_{train}$ 

            \State Construct and sample from a mixture of posterior predictive distributions 
            \State (US) Select $\argmaxA_{c' \in \mathcal{C} \setminus \mathcal{C}_{train}} \!\!\!\!\!\! H(\bs{Y}_{c',:,:\!} | \{\bs{Y}_{c,:,:}\}_{c\in \mathcal{C}_{train}}\!)$ using MC samples 
            \blue \State (IG) Select $\argmaxA_{c' \in \mathcal{C} \setminus \mathcal{C}_{train}} \!\!\!\!\!\! H(\bs{Y}_{c',:,:\!} | \{\bs{Y}_{c,:,:}\}_{c\in \mathcal{C}_{train} }\!)  -\E_{q(\bs{\Theta})}[H(\bs{Y}_{c',:,:} |\bs{\Theta}, \{\bs{Y}_{c,:,:\!}\}_{c\in \mathcal{C}_{train} }\!)] $ 
            \State using MC samples\black
            \State Observe data ($\bs{Y}_{c',:,:}$) under condition $c'$ 
            \State Update $\mathcal{C}_{train} = \mathcal{C}_{train} \cup  \{  {c'} \}$
        \EndWhile
    \end{algorithmic}
\end{algorithm}

At each iteration, 
Algorithm 1 %
fits \texttt{CS-GPFA} model with data from the set of observed conditions $\mathcal{C}_{train}$. 
We seek to evaluate two objectives \textbf{(i)} the \textit{entropy of the posterior predictive distribution} 
$H(\bs{Y}_{c',:,:} | \{\bs{Y}_{c,:,:}\}_{c\in \mathcal{C}_{train} })$
at an unseen condition $c'$ 
and \textbf{(ii)} the \textit{information gain} if we obtain samples from the unseen condition $c'$: 
\[H(\bs{Y}_{c',:,:} | \{\bs{Y}_{c,:,:}\}_{c\in \mathcal{C}_{train} }) -\E_{q(\bs{\Theta})}[H(\bs{Y}_{c',:,:} |\bs{\Theta}, \{\bs{Y}_{c,:,:}\}_{c\in \mathcal{C}_{train} })] .\] 
This first objective  quantifies the model's uncertainty of our predictions under unseen conditions, whilst averaging over the effects of all model variables 
($\bs{\Theta}$, $\bs{X}'$). 
The second objective %
quantifies how much this entropy would be reduced if we were to observe data under the unseen condition $c'$. 
Note that the distribution $p(\bs{Y}_{c',:,:} | \{\bs{Y}_{c,:,:}\}_{c\in \mathcal{C}_{train} }) $ is not analytically tractable.
We approximate it as a finite mixture of distributions using Monte Carlo (MC) samples. Each component of this mixture is in chosen class of our likelihood (e.g. negative binomial).

We estimate the entropies by drawing %
samples from the finite mixture distribution and averaging its log-likelihood under the component distributions (see details in \ref{appendix:active-learning}). 
Our algorithm then greedily selects the condition $c'$ that maximizes the selected objective and observes samples under the condition. 
In our experiments, the last step is simulated by 
adding trials of data for the selected condition.

%% file: section__experiments.tex
\section{Experiments}
\label{sec:experiments}
We evaluate our model on a visual coding dataset and demonstrate its use in adaptive selection of experimental conditions. In  \ref{sec:appendix:experiments}, we include synthetic experiments (\ref{subsec:appendix:synthetic_experiment}), experiments with fewer neurons (\ref{subsec:appendix:visual_coding:fewer_neurons}), and experiments on a center-out reaching dataset (\ref{subsec:appendix:mc_maze_experiment}).

\input{section__experiment_allen_varying_contrast}

\input{section__experiment_active_learning}

%% file: section__experiment_allen_varying_contrast.tex
\subsection{Visual coding experiment}
\label{subsection:experiment:visual_coding_experiment}
\textbf{Dataset} In this experiment, we evaluate our model on a session from a visual coding dataset from Allen Brain Observatory  (\texttt{mouse\_id = 744912849, session\_type = functional connectivity}) \citep{allen-data}. This session includes 15-trial recordings V1 area neurons under drifting gratings visual stimuli with 9 varying levels of contrast. Each trial is 500ms long and binned into a total of 49 10-ms bins. We preprocessed the recordings to filter out neurons with $<$5Hz average firing rate. We parameterize our condition space using the contrast levels.

\textbf{Baselines}
For comparison, we consider (1) \citet{nadew2024conditionallyconjugate}'s {ccGPFA}\footnote{https://github.com/yididiyan/ccgpfa}, 
(2) {VBGCP}\footnote{https://github.com/hugosou/vbgcp}, a tensor decomposition method \citep{soulat2021probabilistic}, 
(3) \texttt{CS-GPFA}, our proposed method, 
hereafter referred as \texttt{CS-GPFA(smooth)}, and 
(4) a simpler diagonal version of our proposed \texttt{CS-GPFA} with zero covariance between the condition points $K_{CC}^{\bs{(c)}} = I$ (\nscsgpfa). In practice, we fixed the corresponding lengthscale parameters to 1e-8. For ccGPFA, we fit the model separately for each condition.

\textbf{Setup}
We first investigated the effect of the number of training trials on model performance. 
We fit the models 
with a varying number of training trials, $1-9$. 
For each choice, we randomly split the dataset and repeated the experiments 10 times. 
In each repeat, we held out 5 trials to evaluate test performance of the models. To accommodate VBGCP for comparison, we averaged the inferred factors across trials as it learns a different factor for each trial. %
In addition, we used the same initialization of dispersion parameters based on the empirical mean firing rate. As a stopping criterion, we used a maximum iteration of 1000 and a common threshold on training log-likelihood. 
For the GPFA-based models, we used a pool of $D=10$ latent processes. We initialized their GP  hyperparameters by matching the correlations of a Mat\'{e}rn-3/2 kernel for \csgpfa and an RBF for ccGPFA at a distance of 1 timestep.

\begin{figure}
\centering

    \begin{subfigure}[b]{0.3\textwidth}
        \includegraphics[width=\linewidth]{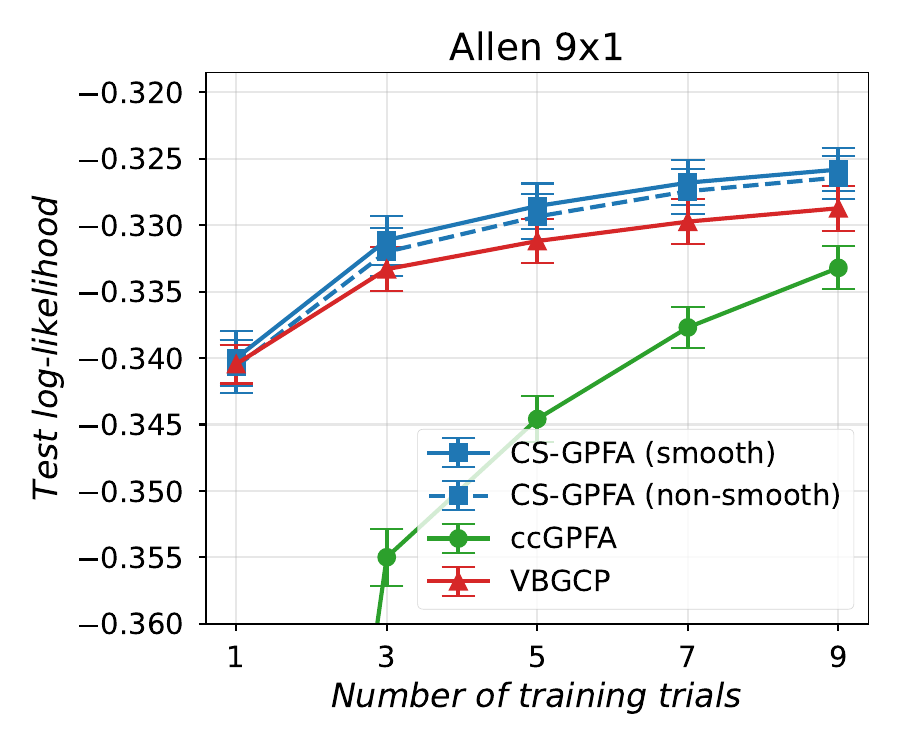}    
        \caption{}
    \end{subfigure}
    \begin{subfigure}[b]{0.3\textwidth}
        \includegraphics[width=\linewidth]{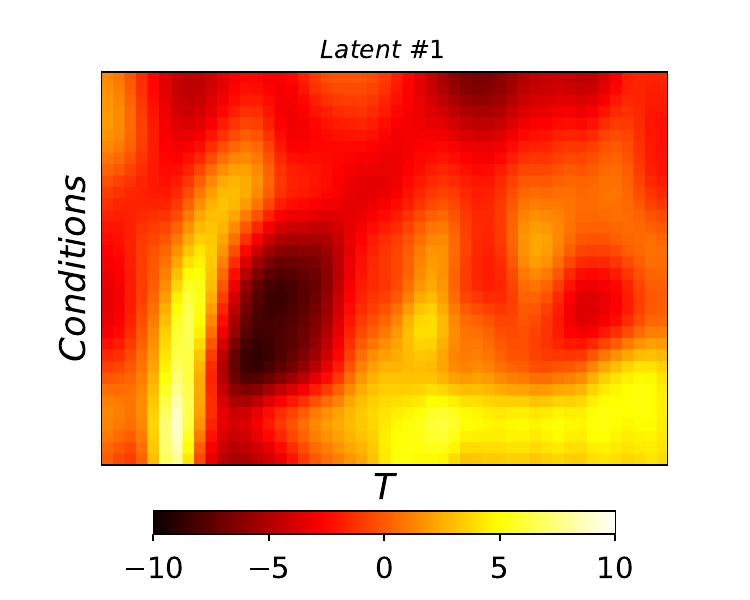}    
        \caption{}
    \end{subfigure}
    \begin{subfigure}[b]{0.35\textwidth}
        \includegraphics[width=\linewidth]{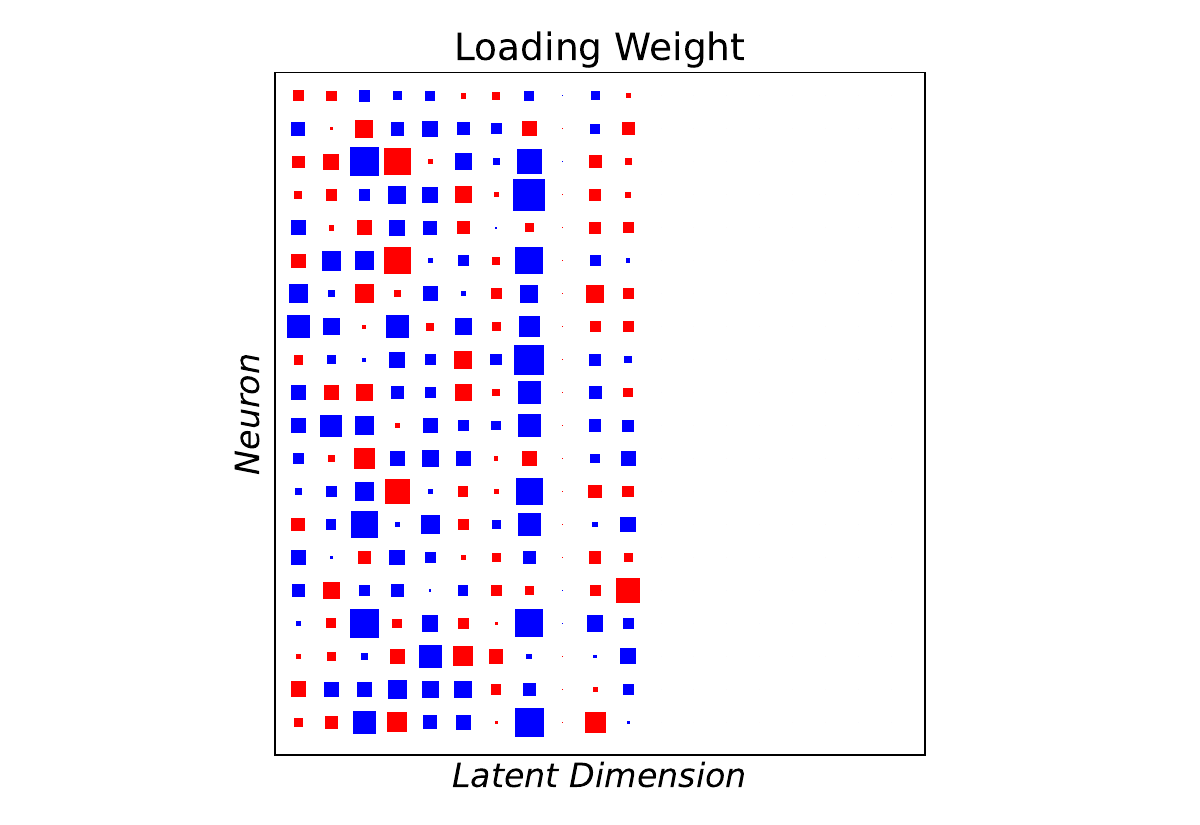}    \caption{}
    \end{subfigure}
    \\
    \caption{\textbf{Visual Coding Experiment: 9 contrast levels} 
    \textbf{(a)} Comparison with baselines using mean test log-likelihood performance across varying numbers of training trials. Both \csgpfa methods outperform the baselines by learning accurate representations.
   \textbf{(b)} The most important latent state (after orthonormalization) across conditions (rows) for our \csgpfa; the last rows (highest contrast levels) show highest latent activity early on, indicating stronger neural response to the stimulus \textbf{(c)} Inferred loading weights for the first 20 neurons (rows) along each latent dimensions (columns); \csgpfa's ARD fully eliminates 9/20 latent dimensions, resulting in a concise representation.
    }
    \label{fig:expts:allen_9x1}
\end{figure}

\textbf{Results} Figure~\ref{fig:expts:allen_9x1}(a) shows the experimental results 
    
with varying training sizes. 
\texttt{CS-GPFA(smooth)} achieves consistent improvement over all baselines 
The improvement over ccGPFA baselines could be explained by the coupled assumption on latent functions, which exploits the structural correlation in the data. 
In addition, the improvement over the \nscsgpfa baseline shows the utility of the smooth priors that directly use the condition labels. The result also shows \csgpfa advantage over VBGCP which does not account for condition differences.

\textbf{Analysis}
In Figure~\ref{fig:expts:allen_9x1}(b), we use our \csgpfa{} method to visualize the inferred latent activity, capturing population dynamics across stimulus conditions. We sampled stimulus conditions with contrast levels in the range [0.01, 1], and then used \csgpfa{} model to predict the corresponding latent activity. To isolate the most informative components, we applied orthonormalization following \citet{yu2008gaussian}.

The resulting trajectories highlight two key patterns. First, population responses increase in magnitude with higher contrast levels (top to bottom) following stimulus onset. Second, the time-to-peak is modulated by contrast: high-contrast stimuli elicit rapid peaks, whereas low-contrast stimuli evoke slower, more gradual responses. These observations are consistent with prior findings on condition-dependent response modulation in the visual cortex \citep[Figure 4]{gebodh2017effects}.

%% file: section__experiment_active_learning.tex
\subsection{Exploring visual coding dataset }

\label{subsec:experiments:active-learning}
\begin{figure}
    \centering
    \begin{subfigure}[b]{0.4\textwidth}
        \includegraphics[width=1.\linewidth]{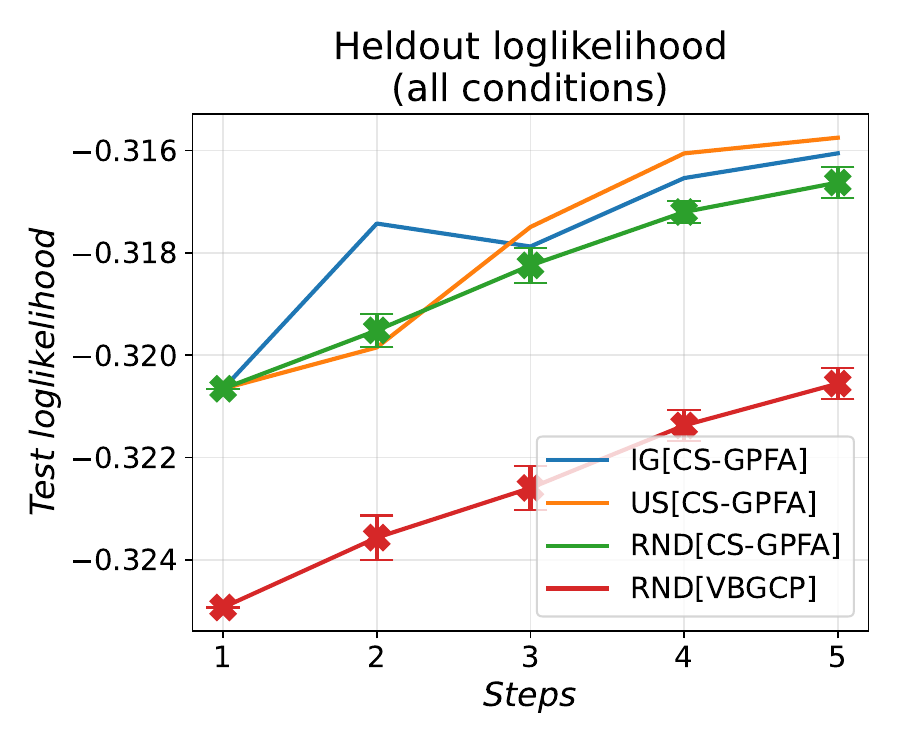}
        \caption{}
    \end{subfigure}
    \begin{subfigure}[b]{0.8\textwidth}
        \includegraphics[width=1.\linewidth]{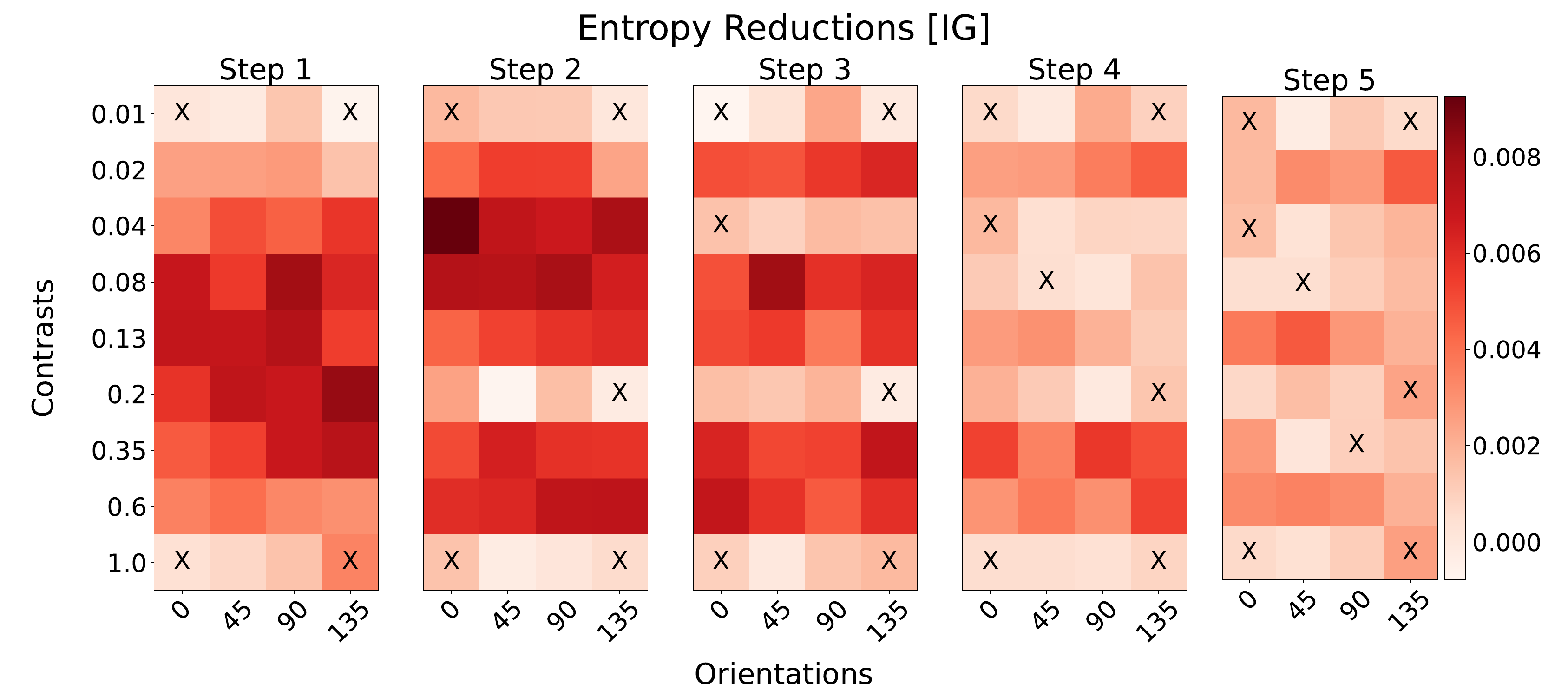}
        \caption{}
    \end{subfigure}
    \begin{subfigure}[b]{0.8\textwidth}
        \includegraphics[width=1.\linewidth]{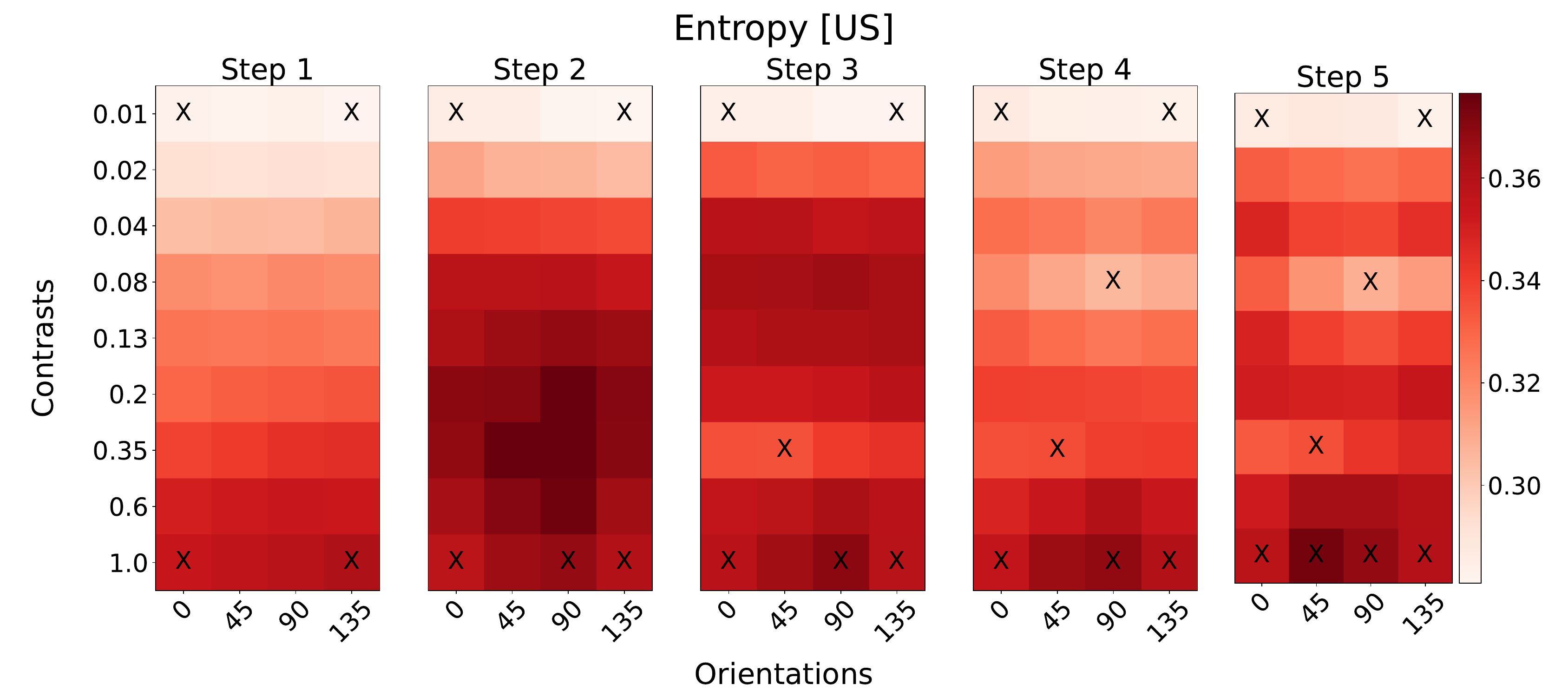}
        \caption{}
    \end{subfigure}
    \caption{\textbf{Active Learning Experiment} 
        \textbf{(a)} Mean held-out log-likelihood estimates at each step. %
        \textbf{(b)} Heatmaps for predictive entropy reductions with \textit{information gain} (IG[ \csgpfa]) objective; `X' indicates the conditions selected at the step
        \textbf{(c)} and heatmaps for predictive entropy with \textit{uncertainty sampling} (US[ \csgpfa]) objective at each step of the algorithm. 
    }
    \label{fig:result:active-learning-1}
\end{figure}

\textbf{Dataset} In this experiment, we consider a ``drifting gratings contrast” session from the Allen Brain Observatory (\texttt{mouse id = 744912849, session type = functional connectivity}) \citep{allen-data}. The session comprises recordings of the V1 area under drifting gratings stimuli defined by two continuous variables: orientation (0°, 45°, 90°, 135°) and contrast (0.01–1). This yields 36 stimuli (9 contrast levels × 4 orientations), forming a 2-D space of experimental conditions.

There are a total of 15 trials per stimulus. 
We preprocessed the trials to filter out neurons with $<5 $ Hz average spike count and binned the spikes into 10 ms windows. 
We then randomly shuffled and split trials into held-in and held-out in a 3:1 ratio. 

\textbf{Setup}
To run our active learning procedure, we initialized the model to be the corner conditions (lowest/highest contrast levels, 0°/135° orientations).
We ran Algorithm 1 with the entropy (US)
and information gain (IG) objectives. 
At each step, we estimated the objectives for all conditions using 100 Monte Carlo samples from the \texttt{CS-GPFA} model. We also computed log-likelihood performance via the held-out trials. For unseen conditions, we use \eqref{eq:latent-prediction} to predict their latent trajectories and corresponding spike probabilities.  

\textbf{Metric} To evaluate the performance of our active learning algorithm, we estimate the mean log-likelihood using the held-out trials under all conditions. In addition, for each step in the algorithm, we report the predictive entropy estimates for all conditions as heatmaps in Figure~\ref{fig:result:active-learning-1}. 

\textbf{Baseline}
We consider two baselines:  \texttt{CS-GPFA} and VBGCP models with random selection of next conditions and model. For VBGCP, since it does not inherently allow for prediction, we used nearest-neighbor to compute log-likelihood on the held-out trials. 

\textbf{Results } Overall, Figure \ref{fig:result:active-learning-1}(a) shows that the active learning method based on \csgpfa outperforms VBGCP across the evaluated steps. This pattern highlights the strong modeling and interpolation capabilities of \csgpfa, particularly due to its explicit treatment of conditions within the latent structure. Among the \csgpfa-based acquisition objectives, both US and IG consistently achieve better performance than the RND baseline. While their advantages over RND are less stable in the initial steps (1–2)—likely due to limited data and less reliable early-stage modeling—by steps 3–5 both strategies demonstrate clear and consistent improvements relative to the random selection strategy.

In Figure~\ref{fig:result:active-learning-1}(b–c), the heatmaps illustrate the points selected by the IG and US acquisition functions under the \csgpfa model. IG selects points that are more broadly distributed across the condition space, reflecting its focus on global uncertainty reduction. In contrast, US concentrates selections at high-contrast conditions, aligning with the higher variability in neural activity observed at these contrast levels.

%% file: appendix__related_works.tex
\section{Related Works - Extended Discussion}
\label{apndx:related-works-extended}

\textbf{Data Augmentation}
\label{sec:related-works:data-augmentation}
While a GPFA model with Negative-Binomial likelihood is well-suited for count data, Bayesian inference of model variables ($ \{ \bs{X}, \bs{W}, \bs{\beta}, \bs{r} \}$) becomes intractable.
Specifically, there is  no known prior for which the (joint/conditional)-posterior distributions are available in closed form. 
Some works %
addressed this problem by using polynomial approximations \citep{keeley2020efficient} or Monte Carlo approximations \citep{jensen2021scalable}.  
\cite{nadew2024conditionallyconjugate} leveraged a data augmentation approach to restore conditional conjugacy to the model and derive closed form variational updates. 
In this work, we extend this framework to a more complex GPFA-based model to derive closed-form updates for it (\ref{sec:appendix:variational-updates}).

\textbf{Deep learning based Methods for Neural Data} \citep{pandarinath2018inferring} use RNN based variational autoencoders to infer latent trajectories from spike data. Similarly, \cite{dowling2024exponential} extends the idea of variational autoencoders for deep state-space models. However, due to their neural network backbone, such methods are not ideal for low-sample regime. In this work, we propose a probabilistic and sample-efficient multi-condition neural model.

\textbf{Other Methods for Neural Data} \citet{nejatbakhsh2023estimating} work on estimating ``noises" in neural populations across continuous conditions. Their method exploits parameterized conditions to learn neuron-to-neuron noise correlations, which is different from the temporal latent representation we discuss here (i.e., they do not model activity over time).

\textbf{Active Learning Methods for Neural Data} Previous studies have investigated adaptive stimulus selection and active learning across different modeling frameworks. \cite{lewi2007efficient} developed a supervised active learning framework for generalized linear models (GLMs), and \cite{park2011active} introduced an uncertainty sampling strategy that aggregates information across trials for Gaussian process (GP)–based regression models. \cite{jha2024active} also proposed an information-theoretic approach for adaptive stimulus selection in a latent variable model. However, this method is restricted to regression models with discrete latent variables. 
In contrast, our work presents the first application of active learning to GPFA-based latent variable models that explicitly preserves the temporal structure of the data.

\textbf{State-space Gaussian Processes} Scalable and exact inference of Gaussian processes has been a longstanding problem of in the literature. For specific stationary Gaussian process kernels that satisfy certain properties in their spectral densities, previous works \citep{hartikainen2010kalman, sarkka2013spatiotemporal, solin2016stochastic, wilkinson2020} have derived an equivalent state-space representation for Gaussian processes with these kernels. This allows for fast linear-time inference via standard Kalman filtering \citep{kalman1960new} and smoothing techniques \citep{rauch1965maximum}. 
Numerous methods, including some works in computational neuroscience literature \citep{dowling2023linear, li2024markovian, li2025learning} 
have applied this formulation to speed up inference in their respective models. 
However, these methods are either restricted to Gaussian observation models and/or offer non-Bayesian treatment to important model variables such as loading weights. In our work, we provide a novel framework that handles both non-conjugacy and Bayesian treatment of all model variables simultaneously.

%% file: appendix__data_augmentation.tex
\section{Data Augmentation}

\subsection{Augmentation}

\label{sec:appendix:augmentation:augmentation}
In this section, we show in detail how we augment auxiliary variables into our model. Recall the likelihood of the spike count under a negative binomial model Eqn. \eqref{eq:negative-binomial-likelhood}. 

\begin{align}
    p(\{ \bs{Y}_c \}_{c=1}^C | \bs{\Theta}) = \prod_{c,n,t}  \bigg ( \frac{\Gamma(\cnt{Y} + \bs{r_n}) }{\cnt{Y}!\Gamma(\bs{r_n})} \bigg )  \bigg (  \frac{ ( e^{\cnt{F}} )^{\cnt{Y}} }{ (1 + e^{\cnt{F}} )^{ \cnt{Y} + \bs{r_n}}  }  \bigg ) \label{eq:appendix:negbinomial-likelihood}
\end{align}

To conduct Bayesian inference on our model variables, $\bs{\Theta}$, we must integrate them out from our model. %
Due to the difficulty of working with integration involving the nonlinear terms analytically, Bayesian inference becomes intractable. 
\citet{polson2013bayesian} proposed 
    
an integral identity to transform the exponential term resulting in an equivalent Gaussian likelihood over a transformed variable. %
\citet{nadew2024conditionallyconjugate}  applied this integral identity on a GPFA model. We follow their approach to make our model \texttt{CS-GPFA} conditionally-conjugate,
yielding tractable Bayesian inference.

Applying an %
integral identity from \cite{polson2013bayesian}, we transform the exponential term as an integral over a new Polya-Gamma auxiliary variable $\cnt{\omega}$
\begin{align}
      \frac{( e^{\cnt{F}} )^{\cnt{Y}} }{ (1 + e^{\cnt{F}} )^{\cnt{Y} + \bs{r}_n } }  &= 2^{-b} \exp((a- b/2) {\cnt{F}} ) \nonumber \\ 
      &\qquad \times \int_0^\infty \exp \big \{ - \frac{1}{2} \cnt{\omega} {\cnt{F}}^2 \big \} \PG(\cnt{\omega}| b, 0) d \cnt{\omega} ,
      \label{eq:appendix:PG-integral-identity}
\end{align}

where $a=\cnt{Y}$ and $b=\cnt{Y} + r_n$.
Simplifying the expression by removing terms with only $\cnt{Y}$ dependence,
     
we get 
\begin{align}
     \frac{( e^{\cnt{F}} )^{\cnt{Y}} }{ (1 + e^{\cnt{F}} )^{\cnt{Y} + \bs{r}_n } }  &= \exp(- \bs{r}_n \log 2  + (\frac{\cnt{Y} - \bs{r}_n}{2}) {\cnt{F}} ) \nonumber\\
     & \times\int_0^\infty \exp \big \{ - \frac{1}{2}\cnt{\omega} {\cnt{F}}^2 \big \} \PG(\cnt{\omega}| \cnt{Y} + \bs{r}_n, 0) d \cnt{\omega} .
    \label{eq:appendix:augmentation:polya-gamma}
\end{align}

The integral identity can be considered as a  marginalization of the auxiliary variables $\hat{\bs{\Theta}} = \{  \cnt{\omega}\}$ from an augmented likelihood,  $p(\bs{Y}, \hat{\bs{\Theta}} | \bs{\Theta})$,  with 
\begin{align*}
    p(\bs{Y} | \bs{\Theta}) = \int_{\hat{\bs{\Theta}}} p(\bs{Y}, \hat{\bs{\Theta}} | \bs{\Theta}) d\hat{\bs{\Theta}}. 
\end{align*}
Therefore, we identify the augmented likelihood as the product of terms inside the integrals. %
\begin{align}
    p(\bs{Y}, \hat{\bs{\Theta}} | \bs{\Theta}) &= \prod_{c,n,t}  \bigg ( \frac{\Gamma(\cnt{Y} + \bs{r_n}) }{\cnt{Y}!\Gamma(\bs{r_n})}  \nonumber \\
    &\qquad \qquad\exp \left (- \bs{r}_n \log 2  + \frac{1}{2}(\cnt{Y} - \bs{r}_n) {\cnt{F}}  - \frac{1}{2}\cnt{\omega} {\cnt{F}}^2  \right ) \nonumber \\
    &\qquad \qquad \times \PG(\cnt{\omega}| \cnt{Y} + \bs{r}_n, 0) \bigg ) . \label{eq:appendix:augmented-likelihood:prodterms} 
\end{align}

Now, we express the above augmented likelihood with respect to our model variables $\cnt{F}$ (dependent on $\{ \bs{X}, \bs{W} \}$) and $\bs{r}_n$.  Expressing \eqref{eq:appendix:augmented-likelihood:prodterms} 
as a function of $\{ \cnt{F} \}$, 

\begin{align}
    p_{\{ \cnt{F} \} }(\bs{Y}, \hat{\bs{\Theta}} | \bs{\Theta})%
    & \propto \prod_{c,n,t} \exp( \frac{1}{2}(\cnt{Y} - \bs{r}_n) {\cnt{F}}  - \frac{1}{2}\cnt{\omega} {\cnt{F}}^2  )  \nonumber \\
    &\propto \prod_{c,n,t} \exp( - \frac{1}{2} \cnt{\omega} \bigg ({ {\cnt{F}}^2} - (\frac{\cnt{Y} - \bs{r}_n}{ \cnt{\omega}}) {\cnt{F}} \bigg ) )  \tag{factoring out $\frac{1}{2}\cnt{\omega}$} \nonumber \\
    &\propto \prod_{c,n,t} \exp( - \frac{1}{2} \cnt{\omega} \bigg ({ {\cnt{F}}^2} - (\frac{\cnt{Y} - \bs{r}_n}{ \cnt{\omega}}) {\cnt{F}} \nonumber \\ 
    & \qquad + \bigg (\frac{\cnt{Y} - \bs{r}_n}{ 2\cnt{\omega}} \bigg )^2   - \bigg (\frac{\cnt{Y} - \bs{r}_n}{ 2\cnt{\omega}} \bigg )^2   \bigg ) ) \tag{completing the square} \\
    &\propto \prod_{c,n,t} \exp( - \frac{1}{2} \cnt{\omega} \bigg ({ {\cnt{F}}} - (\frac{\cnt{Y} - \bs{r}_n}{ 2\cnt{\omega}}) \bigg)^2   -  \frac{1}{2} \cnt{\omega} \bigg (\frac{\cnt{Y} - \bs{r}_n}{ 2\cnt{\omega}} \bigg )^2   ) \nonumber  \\
    &\propto \prod_{c,n,t} \exp( - \frac{1}{2} \cnt{\omega} \bigg (  (\frac{\cnt{Y} - \bs{r}_n}{ 2\cnt{\omega}}) -{\cnt{F}}  \bigg)^2  ), %
\end{align}
where in the last line we removed constants w.r.t. $\cnt{F}$ and used $(a-b)^2 = (b-a)^2$.
Recognizing the above as an unnormalized Gaussian density over a transformed variable $\hat{\bs{Y}}_{c,n,t} := \frac{\cnt{Y}-\bs{r}_n}{2\cnt{\omega}}$, we have
\begin{align}
    p_{\{ \cnt{F} \} }(\bs{Y}, \hat{\bs{\Theta}} | \bs{\Theta}) %
    & \propto 
 \prod_{c,n,t} \mathcal{N} (\hat{\bs{Y}}_{c,n,t} | \cnt{F}, 1/\cnt{\omega}). \label{eq:appendix:augmented-likelihood:F} %
\end{align}

One consequence of Eqn. \eqref{eq:appendix:augmented-likelihood:F} is that as a function $\cnt{F}$, the augmented likelihood is a conjugate likelihood for Gaussian priors on $\bs{X}$ and  $\bs{W}$  in the model, restoring conditional conjugacy to the model.
Combining the prior distributions over our model variables $\bs{\Theta}$ (defined in Section~\ref{sec:model})  with the augmented likelihood in Eqn. \eqref{eq:appendix:augmented-likelihood:prodterms}, the joint distribution over the spike counts and all variables $p(\bs{Y}, \hat{\bs{\Theta}}, \bs{\Theta})$ is 
\begin{align}
    p(\bs{Y}, \hat{\bs{\Theta}}, \bs{\Theta}) &=  p(\bs{\Theta}) p(\bs{Y}, \hat{\bs{\Theta}}| \bs{\Theta}) \nonumber \\
    &=p(\bs{\tau}) p(\bs{W}| \bs{\tau}) p(\bs{X}) \prod_{c,n,t}  \bigg ( \frac{\Gamma(\cnt{Y} + \bs{r_n}) }{\cnt{Y}!\Gamma(\bs{r_n})}\nonumber \\
    &\qquad \times \exp(- \bs{r}_n \log 2  + \frac{1}{2}(\cnt{Y} - \bs{r}_n) {\cnt{F}}  - \frac{1}{2}\cnt{\omega} {\cnt{F}}^2 ) \nonumber \\ 
    & \qquad \times \PG(\cnt{\omega}| \cnt{Y} + \bs{r}_n, 0) \bigg ).
\end{align}

\subsection{Binomial likelihood}
\label{sec:appendix:augmentation:binomial}
In this subsection, we show that the augmentation discussed above also holds for a \texttt{CS-GPFA} model with binomial likelihood. 

Consider the following binomial likelihood model, where $\bs{k}_n$ \ represents the total number of Bernoulli experiments and $\hat{p}_{c,n,t}$ is the probability of success, linked to the latent function $\cnt{F}$ via log-odds.    
\begin{align}
    p(\{ \bs{Y}^{(c)} \}_{c=1}^C | \bs{\Theta}) & \propto \prod_{c,n,t}  {\hat{p}_{c,n,t} }^{\cnt{Y}} (1- \hat{p}_{c,n,t})^{\bs{k}_n - \cnt{Y}} \nonumber \\
    & \propto \prod_{c,n,t} \frac{(e^{\cnt{F}})^{\cnt{Y}} }{ ( 1 + e^\cnt{F} )^{\bs{k}_n}}.
\end{align}

The sigmoidal term in the likelihood can be transformed using the $\PG$ integral identity presented in Eqn. \eqref{eq:appendix:PG-integral-identity},
restoring conditional conjugacy over the model variables $\bs{\Theta}$. 
Following standard practice \citep{keeley2020efficient}, we can set the total number of Bernoulli experiments $\bs{k}_n$  to the maximum number of spikes for the respective neuron $n$.

%% file: appendix__state_space_gaussian_processes.tex
\section{State-Space Gaussian Process}
\label{sec:appendix:state-space-gp-formulation} 

Stationary kernels with rational spectral densities such as Mat\'{e}rn kernels are shown to have equivalent exact state-space formulations \citep{solin2016stochastic}.

\subsection{Mat\'{e}rn kernels}
Mat\'{e}rn kernels form a popular class of stationary covariance kernels. For a temporal function $f(t)$, the covariance between two points $\Cov(f(t), f(t'))$ is given by,

\begin{align}
    \Cov(f(t), f(t')) = {\displaystyle k_{\nu }(|t-t'|)=\sigma ^{2}{\frac {2^{1-\nu }}{\Gamma (\nu )}}{{\Bigg (}{\sqrt {2\nu }}{\frac {|t-t'|}{l }}{\Bigg )}}^{\nu }K_{\nu }{\Bigg (}{\sqrt {2\nu }}{\frac {|t-t'|}{l }}{\Bigg )},}
\end{align}
where $\sigma^2$ is the magnitude scale parameter, $\nu$ the smoothness parameter, 
$\Gamma$ the gamma function, and 
$K_\nu$ the modified Bessel function of the second kind \citep{solin2016stochastic}. Common half-integer choices for the smoothness parameter $1/2, 3/2, 5/2$ result in Mat\'{e}rn-$1/2$, Mat\'{e}rn-$3/2$ and Mat\'{e}rn-$5/2$ respectively. For a Mat\'{e}rn kernel with smoothness $\nu$, its corresponding Gaussian process, $f$, is $\lfloor \nu \rfloor$ times differentiable.  

For $\nu=3/2$, the kernel function simplifies as,

\begin{align}
    {\displaystyle k_{3/2}(|t-t'|)=\sigma^{2}{\Bigg (}{1 + \sqrt {3}}{\frac {|t-t'|}{l }}{\Bigg )} \exp{\Bigg (}{- \sqrt {3}}{\frac {|t-t'|}{l }}{\Bigg )}
    }
\end{align}

For all half-integer choices including this kernel, all the quantities that determine its state-space dynamics are available in closed form \citep{solin2016stochastic}. For instance for Mat\'{e}rn-3/2,

\begin{itemize}
    \item  \textit{State representation}  $\bs{f} = [f, \frac{\partial{f}}{\partial t}]^\top$
    \item \textit{Transition matrix}  $\bs{A}_t= e^{-\lambda \Delta t} \begin{bmatrix}
    1+\lambda\Delta t & \Delta t\\ 
    \lambda^2\Delta t & 1 - \lambda \Delta t
\end{bmatrix}$, where $\lambda = \sqrt{3}/l$ and $\Delta t$ is the corresponding gap between the $t$-th and $(t+1)$-th observations. 

    \item \textit{Stationary distribution} ${\bs{P}_t} = \begin{bmatrix}
        \sigma^2 & 0 \\
        0 & 3 \sigma^2 / l^2
    \end{bmatrix}  $
    \item \textit{Noise covariance} $\bs{Q_t} = {\bs{P}_t} - \bs{A}_t {\bs{P}_t} \bs{A}_t^\top$, 
    \item \textit{Measurement model} $\bs{H_t} = [1, 0]$ isolates the function $f$ from the state.
\end{itemize}

We conducted all experiments using Mat\'{e}rn-3/2 kernel. We used a fixed scale parameter $\sigma^2$ to 1 to avoid identifiability issues with the loading weights.

%% file: appendix__variational_updates.tex
\section{Variational Updates}
\label{sec:appendix:variational-updates}
In this section, we show how we can exploit the form of the augmented likelihood to derive closed-form updates for our variational distributions on both augmented and model variables.

\subsection{Augmented variables }

For the set of augmented PG variables $\hat{\bs{\Theta}}=$ $\{ \cnt{\omega} \}$ (see \ref{sec:appendix:augmentation:augmentation} for details),  
we can derive their optimal variational distributions 
using the mean field equation Eqn. \eqref{eq:var-update-equation}
as follows, 
\begin{align}
    q^*(\cnt{\omega}) &\propto \exp \{ \E_{q(-\cnt{\omega} )} [\log p(\bs{Y}, \hat{\bs{\Theta}}, \bs{\Theta})] \} \nonumber \\
    & \propto  \exp \bigg \{ \E \bigg [ \log \exp( -\frac{\cnt{\omega} {\cnt{F}}^2}{2} ) \PG(\cnt{\omega}| \cnt{Y} + \bs{r}_n, 0)
    \bigg ] \bigg  \}  \tag{removing terms with no $\cnt{\omega}$ dependence } \\
    & \propto  \exp \bigg \{ \E \bigg [  -\frac{\cnt{\omega} {\cnt{F}}^2}{2}  + \log  \PG(\cnt{\omega}| \cnt{Y} + \bs{r}_n, 0)
    \bigg ] \bigg  \} \nonumber \\ %
    & \propto  \exp \bigg \{  -\frac{\cnt{\omega} \E[{\cnt{F}}^2]}{2}  + \log  \PG(\cnt{\omega}| \cnt{Y} + \bs{r}_n, 0) \bigg  \}  \tag{applying expectation  } \\
     & \propto  \exp \bigg \{ \log \exp( -\frac{\cnt{\omega} \E[{\cnt{F}}^2]}{2} ) \PG(\cnt{\omega}| \cnt{Y} + \bs{r}_n, 0)
     \bigg  \}  \tag{rearranging } \\
     & \propto  \exp \bigg \{ \log \PG(\cnt{\omega}| \cnt{Y} + \bs{r}_n, \sqrt{\E[{\cnt{F}}^2]})
     \bigg  \}  \tag{exponential tilting of PG distribution (up to a normalization constant) } \\
    & \propto  \PG(\cnt{\omega}| \cnt{Y} + \bs{r}_n, \sqrt{\E[{\cnt{F}}^2]}).
    \label{eq:var-update:omega}
\end{align}
Therefore, we get
\begin{align}
    q^*(\cnt{\omega} )& \propto  \PG(\cnt{\omega}| \cnt{Y} + \bs{r}_n, \sqrt{\E[{\cnt{F}}^2]}) .
\end{align}
Note that the augmented variables $\{ \cnt{\omega} \}$ do not have interdependence in their updates, hence their updates can be done in parallel. %

\subsection{Weights}

    \begin{align}
    q^*&(\bs{W}) \nonumber\\
        &\propto \exp \{ \E_{q(-\bs{W})} [\log p(\bs{Y}, \hat{\bs{\Theta}}, \bs{\Theta} )]  \} \nonumber \\
        &\propto \exp \{ \E [ \log p(\bs{Y}, \hat{\bs{\Theta}}| \bs{\Theta} ) p(\bs{W} | \bs{\tau})]  \} \tag{removing terms with no $\bs{W}$ dependence} \nonumber \\
        &\propto \exp \{ \E [ \sum_n \log p(\bs{Y}_{:, n, :}, \hat{\bs{\Theta}}| \bs{\Theta} ) p(\bs{W}_{n, :} | \bs{\tau})]  \}  \nonumber \\
        &\propto \prod_n \exp(\E[ \log p(\bs{W}_{n, :} | \bs{\tau}) ] ) \exp \{ \E [ \log p(\bs{Y}_{:, n, :}, \hat{\bs{\Theta}}| \bs{\Theta} ) ]  \}  \nonumber \\
        &\propto \prod_n \exp(\E[ \log p(\bs{W}_{n, :} | \bs{\tau}) ] ) \exp \{ -\frac{1}{2} \E \bigg [\sum_{c,t}  - 2 \cnt{F} \cnt{\omega}  \hat{\bs{Y}}_{c,n,t}  + \cnt{\omega}  {\cnt{F}}^2 \bigg ]  \} \tag{removing constants w.r.t $\bs{W}$}\nonumber \\  \nonumber \\
        \intertext{
        Here $\cnt{F} = \sum_d \bs{W}_{n,d} \bs{X}_{c,d,t} = \bs{W}_{n,:} \bs{X}_{c,:,t} $. And $\E[\cnt{F}] = \bs{W}_{n,:} \E[\bs{X}_{c,:,t}] =  \E[\bs{X}_{c,:,t}]^\top \bs{W}_{n,:}^\top$. And the quadratic expectation $\E[\cnt{F}^2] = \bs{W}_{n,:} \E[\bs{X}_{c,:,t} \bs{X}_{c,:,t}]^\top]  \bs{W}_{n,:}^\top $. Applying the expectation in the above expression, substituting expectations and applying summation w.r.t. $t$ and $c$, we get
        }
        &\propto \prod_n \exp(\E[ \log p(\bs{W}_{n, :} | \bs{\tau}) ] ) \exp \{\bigg ( \sum_{c,t} \E[\cnt{\omega}  \hat{\bs{Y}}_{c,n,t}] \E[\bs{X}_{c,:,t}]^\top \bigg ) \bs{W}_{n,:}^\top \nonumber  \\
        & \qquad \qquad - \frac{1}{2} \bs{W}_{n,:} \bigg ( \sum_{c,t} \E[\cnt{\omega}]\E[\bs{X}_{c,:,t} \bs{X}_{c,:,t}]^\top]  \bigg )\bs{W}_{n,:}^\top   \}   \nonumber \\
        &\propto \prod_n  \exp \{    \bigg ( \sum_{c,t} \E[\cnt{\omega}  \hat{\bs{Y}}_{c,n,t}] \E[\bs{X}_{c,:,t}]^\top \bigg ) \bs{W}_{n,:}^\top  \nonumber \\
        & \qquad \qquad - \frac{1}{2} \bs{W}_{n,:} \bigg ( \diag(\bs{\tau})  + \sum_{c,t} \E[\cnt{\omega}]\E[\bs{X}_{c,:,t} \bs{X}_{c,:,t}]^\top]  \bigg )\bs{W}_{n,:}^\top \}. 
        \tag{definition of $p(\bs{W}_{n, :} | \bs{\tau})$; applying expectation w.r.t. $q(\bs{\tau})$} \nonumber %
    \end{align}

    We can recognize the above expression as a product of multivariate Gaussian distributions over random vectors $\bs{W}_{n,:}$ with mean $\bs{\mu}_{n}$ and variance $\bs{V}_{n}$,
    \begin{align}
        q^*(\bs{W}) = \prod_n \mathcal{N}(\bs{W}_{n,:}|\bs{m}_{n}, \bs{V}_{n}) 
        \nonumber %
    \end{align}
    where 
    \begin{align}
        \bs{m}_{n} &= \bs{V}_{n} ( \sum_{c,t} \E[\cnt{\omega}  \hat{\bs{Y}}_{c,n,t}] \E[\bs{X}_{c,:,t}]^\top \bigg )^\top \nonumber \\
        \bs{V}_{n} &= \bigg ( \diag(\bs{\tau})  + \sum_{c,t} \E[\cnt{\omega}]\E[\bs{X}_{c,:,t} \bs{X}_{c,:,t}]^\top]  \bigg )
    \end{align}
    
    The variational expectation 
    $\E [\cnt{\omega}   \hat{\bs{Y}}_{c,n,t} ]$ is simply $ \frac{\cnt{Y} - \bs{r}_n }{2}$ since $\hat{\bs{Y}}_{c,n,t}  =  \frac{\cnt{Y} - \bs{r}_n }{2 \cnt{\omega}} $. 
\black

\subsection{Latent processes}

For the first $D - 1$ latent processes, we update their distribution as follows. %
For the $D$-th latent process, corresponding to the bias terms, we keep the mean to exactly 1 and the variance set to a very small value to induce a constant GP function. 

Here, we first simplify the expression of the optimal latent processes $q^*(\bs{X}_{:,d,:})$ using the mean field equation and invoke the state-space equivalent formulation (See \ref{sec:appendix:state-space-gp-formulation} for details) by rewriting $p^*(\bs{X}_{:,d,:})$ as marginalized from of joint augmented distribution $p(\overline{\bs{X}}_{:,d,:})$.
    {\allowdisplaybreaks
    \begin{align}
        q^*&(\bs{X}_{:,d,:}) \nonumber\\
        &\propto p(\bs{X}_{:,d,:}) \exp \{ \E_{q(-\bs{X}_{d})} [\log p(\bs{Y},\hat{\bs{\Theta}} | \bs{\Theta} )]  \} \nonumber \\
        &\propto p(\bs{X}_{:,d,:}) \exp \{ -\frac{1}{2} \E \bigg [\sum_{c,n,t} \cnt{\omega} (\hat{\bs{Y}}_{c,n,t} - \cnt{F})^2   \bigg ]  \} \nonumber \\
        &\propto p(\bs{X}_{:,d,:}) \exp \{ -\frac{1}{2} \E \bigg [\sum_{c,n,t}  - 2 \cnt{\omega}\hat{\bs{Y}}_{c,n,t}\cnt{F}  + \cnt{\omega} \cnt{F}^2   \bigg ]  \} \tag{dropping terms with no $\bs{X}$ dependence} \nonumber \\
        &\propto p(\bs{X}_{:,d,:}) \exp \{ -\frac{1}{2} \E \bigg [\sum_{c,n,t}  - 2 \cnt{\omega}\hat{\bs{Y}}_{c,n,t} \sum_{d'} \bs{W}_{n,d'} \bs{X}_{c, d', t}   + \cnt{\omega} \left ( \sum_{d'} \bs{W}_{n,d'} \bs{X}_{c, d', t} \right)^2   \bigg ]  \} \tag{$\cnt{F} = \sum_{d'} \bs{W}_{n,d'} \bs{X}_{c, d', t}  $} \nonumber \\
        &\propto p(\bs{X}_{:,d,:}) \exp \{ -\frac{1}{2} \E \bigg [\sum_{c,n,t}  - 2 \cnt{\omega}\hat{\bs{Y}}_{c,n,t}  \bs{W}_{n,d} \bs{X}_{c, d, t}   + \cnt{\omega} \left ( \sum_{d'} \bs{W}_{n,d'} \bs{X}_{c, d', t} \right)^2   \bigg ]  \} \tag{dropping terms with no $\bs{X}_{:,d,:}$ dependence} \nonumber \\
        &\propto p(\bs{X}_{:,d,:}) \exp \{ -\frac{1}{2} \E \bigg [\sum_{c,n,t}  - 2 \left (\cnt{\omega}  \hat{\bs{Y}}_{c,n,t}  \bs{W}_{n,d} - \cnt{\omega}  \sum_{d'\neq d} \bs{W}_{n,d} \bs{X}_{c,d',t} \right )\bs{X}_{c, d, t} \nonumber \\
        &\qquad \qquad + \cnt{\omega} \bs{W}_{n,d}^2 \bs{X}_{c,d,t}^2  \bigg ]  \} \tag{dropping terms with no $\bs{X}_{:,d,:}$ dependence} \nonumber \\
        &\propto p(\bs{X}_{:,d,:}) \exp \{ -\frac{1}{2} \E \bigg [\sum_{c,t}  - 2 \left (\sum_n \cnt{\omega}  \hat{\bs{Y}}_{c,n,t}  \bs{W}_{n,d} - \cnt{\omega}  \sum_{d'\neq d} \bs{W}_{n,d} \bs{X}_{c,d',t} \right )\bs{X}_{c, d, t} \nonumber \\
        &\qquad \qquad  + \sum_n \cnt{\omega} \bs{W}_{n,d}^2 \bs{X}_{c,d,t}^2  \bigg ]  \} \tag{applying summation} \nonumber \\
        &\propto p(\bs{X}_{:,d,:}) \exp \{ \sum_{c,t}  \underbrace{\E \bigg [\sum_n \cnt{\omega}  \hat{\bs{Y}}_{c,n,t}  \bs{W}_{n,d} - \cnt{\omega}  \sum_{d'\neq d} \bs{W}_{n,d} \bs{X}_{c,d',t} \bigg ]}_{:= \bs{\Phi}_{c,t}} \bs{X}_{c, d, t} 
        \nonumber \\
        &\qquad \qquad -\frac{1}{2} \underbrace{\E \bigg [\sum_n \cnt{\omega} \bs{W}_{n,d}^2 \bigg]}_{:=\bs{\Psi}_{c,t}} \bs{X}_{c,d,t}^2   \} \tag{applying expectation } \nonumber \\
        &\propto p(\bs{X}_{:,d,:}) \exp \{ \sum_{t}  \begin{bmatrix}
            \bs{\Phi}_{1,t} \\
            \bs{\Phi}_{2,t}\\
            \vdots \\
            \bs{\Phi}_{C,t}
        \end{bmatrix}^\top 
        \bs{X}_{:, d, t} -\frac{1}{2} \bs{X}_{:,d,t}^\top \diag(\bs{\Psi}_{1,t},...,\bs{\Psi}_{C,t}) \bs{X}_{:,d,t}   \} \tag{vectorized notation 
        } \nonumber \\
        &\propto p(\bs{X}_{:,d,:}) \exp \{ \sum_{t} - \frac{1}{2} \bigg ( \diag(\bs{\Psi}_{1,t},...,\bs{\Psi}_{C,t})^{-1} \begin{bmatrix}
            \bs{\Phi}_{1,t} \\
            \bs{\Phi}_{2,t}\\
            \vdots \\
            \bs{\Phi}_{C,t}
        \end{bmatrix}  
        - \bs{X}_{:, d, t}  \bigg )^\top 
        \diag(\bs{\Psi}_{1,t},...,\bs{\Psi}_{C,t})\\
        &\qquad \qquad \bigg ( \diag(\bs{\Psi}_{1,t},...,\bs{\Psi}_{C,t})^{-1} \begin{bmatrix}
            \bs{\Phi}_{1,t} \\
            \bs{\Phi}_{2,t}\\
            \vdots \\
            \bs{\Phi}_{C,t}
        \end{bmatrix}%
        - \bs{X}_{:, d, t}  \bigg )\} \tag{ completing the square} \nonumber \\
    \intertext{We recognize the above as unnormalized Gaussian likelihood of a pseudo observation, $\diag(\bs{\Psi}_{1,t},...,\bs{\Psi}_{C,t})^{-1} \begin{bmatrix}
            \bs{\Phi}_{1,t} \\
            \bs{\Phi}_{2,t}\\
            \vdots \\
            \bs{\Phi}_{C,t}
        \end{bmatrix}$, with mean and noise precision $\bs{X}_{:,d,t}$ and $\diag(\bs{\Psi}_{1,t},...,\bs{\Psi}_{C,t})$ respectively.  Recall $p(\bs{X}_{:,d,:})$ for separable Markovian kernels, can be transformed into its state-space equivalent induced by the corresponding Gaussian process kernel, with $T$ discrete states. Representing $\bs{X}_{:,d,t}$ via projection of its corresponding state, we get $\bs{X}_{:,d,t} = \bs{H} \obs{X}_{:,d,t}$,  the marginal posterior distributions $\{q(\bs{X}_{:,d,t})\}_{t\in [1...T]}$ can be efficiently calculated using state-of-the-art Kalman filter and smoothing algorithms in linear time with respect to the number of states $T$ \citep{sarkka2013spatiotemporal}. Furthermore, we note that in the other variational update rules, computing the  variational expectations w.r.t. $q(\bs{X}_{:,d,:})$ only require its marginal variances. Therefore, we effectively avoid the need to learn the full form of the posterior.}
    \end{align}

    }

\subsection{ARD gamma variables}

\begin{align}
        q^*(\bs{\tau}) &\propto p(\bs{\tau}) \exp \{  \E_{q(-\bs{\tau})} [\log \prod_n p(\bs{W}_{n, :} | \bs{\tau})] \}  \tag{removing terms with no $\bs{\tau}$ dependence}\nonumber \\
        &\propto p(\bs{\tau}) \exp \{  \E [ \sum_n \frac{1}{2} \log |\diag(\bs{\tau})| - \frac{1}{2}  \bs{W}_{n, :} \diag ( \bs{\tau} )   \bs{W}_{n, :}^\top  ] \} \tag{$p(\bs{W}_{n, :})$ (Section~\ref{subsection:method:more-priors}) and removing constants w.r.t. $\bs{\tau}$}\nonumber \\ 
        &\propto p(\bs{\tau}) \exp \{  \E [ \sum_d \frac{N}{2} \log \bs{\tau_d}  - \sum_d \sum_n \frac{1}{2} \bs{W}_{n, d}^2  \bs{\tau}_d   ] \} \tag{$ \log |\diag(\bs{\tau})| = \sum_d \log \bs{\tau_d}$}\nonumber \\
        &\propto \prod_d p(\bs{\tau}_d) \exp \{  \sum_d \frac{N}{2} \log \bs{\tau_d}  - \sum_d \sum_n \frac{1}{2} \E[\bs{W}_{n, d}^2]  \bs{\tau}_d   \} \tag{$p(\bs{\tau}) = \prod_d p(\bs{\tau}_d)$; taking expectation}\nonumber \\
        &\propto \prod_d  \exp \{   ( \alpha_d + \frac{N}{2} - 1) \log \bs{\tau_d}  - (\beta_d + \frac{1}{2} \sum_n  \E[\bs{W}_{n, d}^2])  \bs{\tau}_d   \} \tag{definition of $p(\bs{\tau}_d)$}\nonumber \\
    \end{align}
\black
Recognizing the last expression as a PDF of the product of gamma distributions, we write the optimal distribution as 

\begin{align}
    q^*(\bs{\tau}) = \prod_d \Gamma( \alpha_d + \frac{N}{2}, \beta_d + \frac{1}{2} \sum_n \E[\bs{W}_{n,d}^2]).
\end{align}

\subsection{M-step updates}

Following a cycle of the above variational distribution updates, we optimize parameters such as dispersion parameters and GP lengthscales by directly optimizing using the ELBO objective. For the GP hyperparameters, we used Pytorch implementation of Adam optimizer \citep{kingma2014adam} with a learning rate of 0.01. We optimized the hyperparameters using Scipy's implementation of the derivative-free Nelder-Mead method \citep{gao2012implementing}.

\subsection{Variational moments}
The quadratic expectation of the latent combination {$ \E[\cnt{F}^2]$ } is computed as follows

    \begin{align}
        \E[\cnt{F}^2] &=\E[ \bs{W}_{n,:} \bs{X}_{c,:,t} \bs{X}_{c,:,t}^\top \bs{W}_{n,:}^\top ] \\ 
        &= \E[ \bs{W}_{n,:} \underbrace{\E[\bs{X}_{c,:,t} \bs{X}_{c,:,t}^\top}_{\Phi } ]\bs{W}_{n,:}^\top ] \\ 
        & = \Tr(\Cov(\bs{W}_{n,:}, \bs{W}_{n,:}) \Phi ) + \E[\bs{W}_{n,:}] \Phi \E[\bs{W}_{n,:}^\top]  ]
    \end{align}

    Here $\Phi = \Cov(\bs{X}_{c,:, t}, \bs{X}_{c,:, t}) + \E[\bs{X}_{c,:, t}] \E[\bs{X}_{c,:, t}^\top] $ is a $D \times D$ matrix and  $\Cov(\bs{X}_{c,:, t}, \bs{X}_{c,:, t})$ is a diagonal matrix, since different latent dimensions are assumed independent in our mean-field partitions $\prod_d q(\bs{X}_{:,d,:})$.

%% file: appendix__active_learning.tex
\section{Active Learning}

\label{appendix:active-learning}

\newcommand{\Yprime}{\bs{Y}_{c',:,:}}

\newcommand{\Xprime}{\bs{X}_{c',:,:}}

\newcommand{\Yobs}{ \{\bs{Y}_{c,:,:}\}_{c\in \mathcal{C}_{train} }}

    \subsection{ Deriving the posterior predictive distribution: $p(\bs{Y}_{c',:,:} | \Yobs)$ }
    \label{subsec:appendix:posterior_prediction}
    
    In this subsection, we show how we can compute the posterior predictive distribution for an unseen condition $c'$, denoted as $p(\Yprime |  \Yobs)$. 
    Recall $\bs{\Theta} = \{ \bs{X}, \bs{W}, \bs{\tau}, \{\cnt{\omega} \} \} $ is the set of all variables in the model.

    Denoting the predicted set of latent process for condition $c'$ as $\Xprime$, the posterior predictive distribution can be written as 
    \begin{align}
        p(\Yprime|\Yobs) &= \int p(\Yprime| \bs{\Theta}, \Xprime) p (\Xprime, \bs{\Theta} |  \Yobs ) d \bs{\Theta} d \Xprime  \nonumber \\ 
        &= \int p(\Yprime | \Xprime, \bs{W}) p (\Xprime, \bs{\Theta} |  \Yobs) d \bs{\Theta} d \Xprime  
        \tag{$\Yprime \perp (\bs{\Theta} \setminus \{ \bs{W}  \} ) | \Xprime, \bs{W}$ }.
    \end{align}

    Assuming the set of points $\mathcal{C}_{train}$ are representative,
    the joint distribution factorizes as $p (\Xprime, \bs{\Theta} |  \Yobs) \approx p (\Xprime | \bs{\Theta}) p(\bs{\Theta} |   \Yobs) $. 
    Furthermore, using our variational approximation of the posterior, $p(\bs{\Theta} | \Yobs) \approx q(\bs{\Theta)}$. Plugging this back in, 
    \begin{align*}
        p(\Yprime | \Yobs  ) & \approx \int p(\Yprime | \bs{W}, \Xprime)  p (\Xprime | \bs{\Theta}) q( \bs{\Theta}) d \bs{\Theta} d \Xprime \\
        & = \int p(\Yprime | \bs{W}, \Xprime)  p (\Xprime | \bs{X}) q( \bs{\Theta}) d \bs{\Theta} d \Xprime \tag{independence}.
    \end{align*}
    The above expression is not analytically tractable. Thus, we propose a Monte-Carlo approximation by sampling set of $L$   realizations $\{ \Xprime, \bs{W} \}$ from $ p (\Xprime | \bs{X}) q( \bs{\Theta})$. 
    Denoting each realization with index $l$, we write the approximation as, 

    \begin{align*}
        p(\Yprime| \bs{Y}) \approx \sum_{l=1}^L p(\Yprime| \Theta^{(l)})
        =: p_\mathrm{mix}(\Yprime| \bs{Y}) ,
    \end{align*}
    where $p_\mathrm{mix}$ denotes the resulting mixture distribution. Note that the components of this mixture distribution are from the same class of distributions as the likelihood model. This fact will be important in the estimation of the entropy of the distribution.
    
\black 
    
    \subsection{Estimating the entropy $H[\Yprime| \Yobs  ]$ }
    \label{appendix:active-learning-entropy}

    As discussed in Section~\ref{sec:active-learning}, we wish to maximize the entropy of the posterior predictive conditional on the unseen condition $c'$ at each step.  Mathematically, the objective is written as follows,
    \begin{align}    
        &\hspace{-2cm}
        \argmaxA_{c' \in \mathcal{C} \setminus \mathcal{C}_{train}} H[\Yprime| \Yobs] \nonumber\\
        &\approx \argmaxA_{c' \in \mathcal{C} \setminus \mathcal{C}_{train}} H[p_\mathrm{mix}(\Yprime| \Yobs)] \tag{use $p_\mathrm{mix}$ as surrogate}  \\
        &\approx \argmaxA_{c' \in \mathcal{C} \setminus \mathcal{C}_{train}} -\frac{1}{Z} \sum_{z=1}^{Z} \log \frac{1}{L} \sum_{l=1}^L  p(\hat{y}_z|\Theta_{c'}^{(l)}), 
        \label{eq:active-learning-objective}
    \end{align}
    where we empirically estimate the entropy of $p_\mathrm{mix}$ and where
    $\hat{y}_z \sim p_\mathrm{mix}(\Yprime| \bs{Y})$.

\subsection{Estimating the information gain \\ $H[\Yprime| \Yobs] - \E_{q(\bs{\Theta})}[H(\bs{Y}_{c',:,:} |\bs{\Theta}, \{\bs{Y}_{c,:,:}\}_{c\in \mathcal{C}_{train} })$ }
\label{appendix:active-learning-information-gain}
We similarly estimate the \textit{expected entropy} term in the information gain objective by sampling model variables $\bs{\Theta}_z \sim q(\bs{\Theta)}$ from the posterior distribution and averaging the predictive entropy under each resulting model (also done empirically via Monte Carlo samples).

%% file: appendix__experiments.tex
\section{Additional Experiments}\label{sec:appendix:experiments}

We conducted additional experiments on both a synthetic dataset and two real-world datasets—the visual coding dataset presented in the main paper and a reaching task dataset from \cite{churchland2010cortical}. In particular, we evaluate our model’s performance in terms of held-out log-likelihood, estimated firing rates, and automatic relevance determination (i.e., selecting the appropriate latent dimensionality.) For the visual coding dataset, we further examine how the number of recorded neurons (channels) affects the recovery of the latent structure and compare test performance against baseline methods.

In addition, we provide negative log-likelihood plots for the active learning experiment to complement the results in the main section.

\subsection{Synthetic experiment}
\label{subsec:appendix:synthetic_experiment}

\textbf{Dataset} We generated a synthetic neural dataset with $D=3$ latent processes, $N=30$ neurons, $M=10$ experimental conditions, and
$T=100$ time steps. Each latent process was modeled as a sinusoidal function with randomly sampled amplitudes and phase shifts to introduce smooth temporal structure. We simulated smoothness along the condition axis by again generating a sinusoidal function with $M=10$ points and computing an outer product with $D=3$ latent processes. Neuronal firing rates were obtained by projecting the latent processes through neuron-specific weight vectors sampled from a standard normal distribution. Spike counts were then generated from our negative binomial observation model, with neuron-specific dispersion parameters drawn uniformly from Uniform[1,5]. For each condition, we simulated 15 independent trials (10 held-in and 5 held-out). 

\textbf{Setup} We first fit \csgpfa with varying number of latent dimensionality $\{ 1, 2, 3, 4, 9, 10\}$ using the held-in trials and evaluated the log-likelihood performance on the held-out trials. In addition, to explore the performance of the model with other baselines (\nscsgpfa, ccGPFA, and VBGCP) in low-data settings, we used $\{ 1, 3, 5, 7, 9\}$ held-in trials to train the models and evaluate their log-likelihood performance on the held-out set and reconstruction of the true firing rate measure using the R2 metric.    

\subsubsection{Results}
\begin{figure}
    \begin{subfigure}[b]{0.4\textwidth}
        \includegraphics[width=\textwidth]{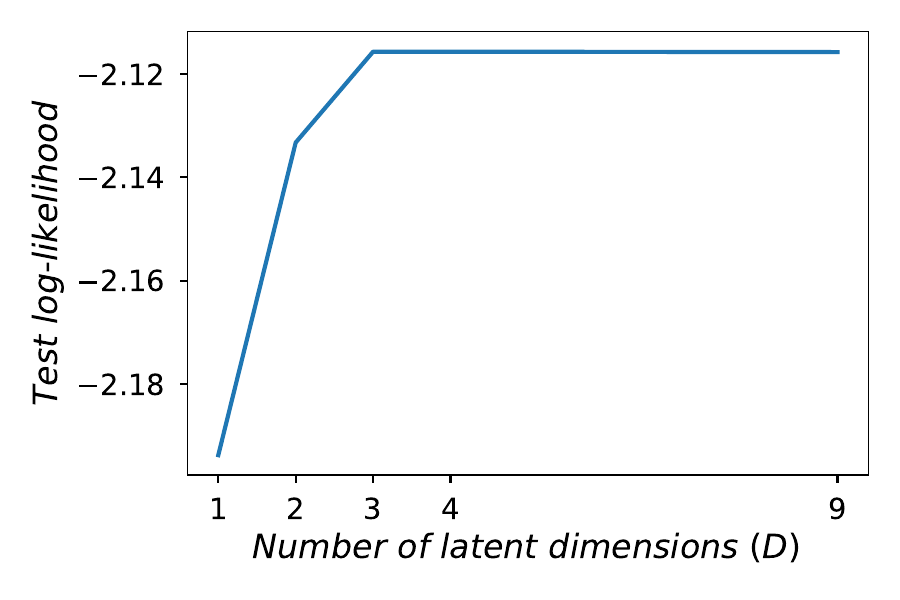}
        \caption{Effect of varying  the latent dimensionality}
    \end{subfigure}
    \hfill
    \begin{subfigure}[b]{0.3\textwidth}
      \includegraphics[width=\textwidth]{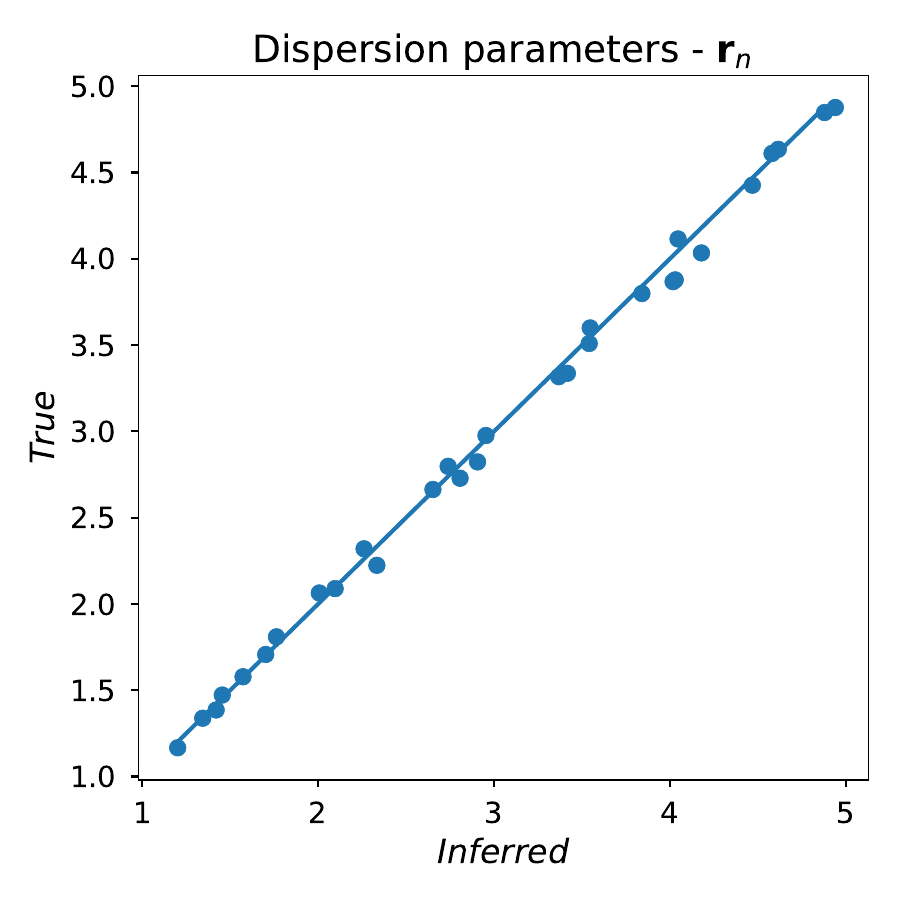}
      \caption{Inferred dispersion parameters}
    \end{subfigure} \\
    \begin{subfigure}[b]{0.4\textwidth}
        \includegraphics[width=\textwidth]{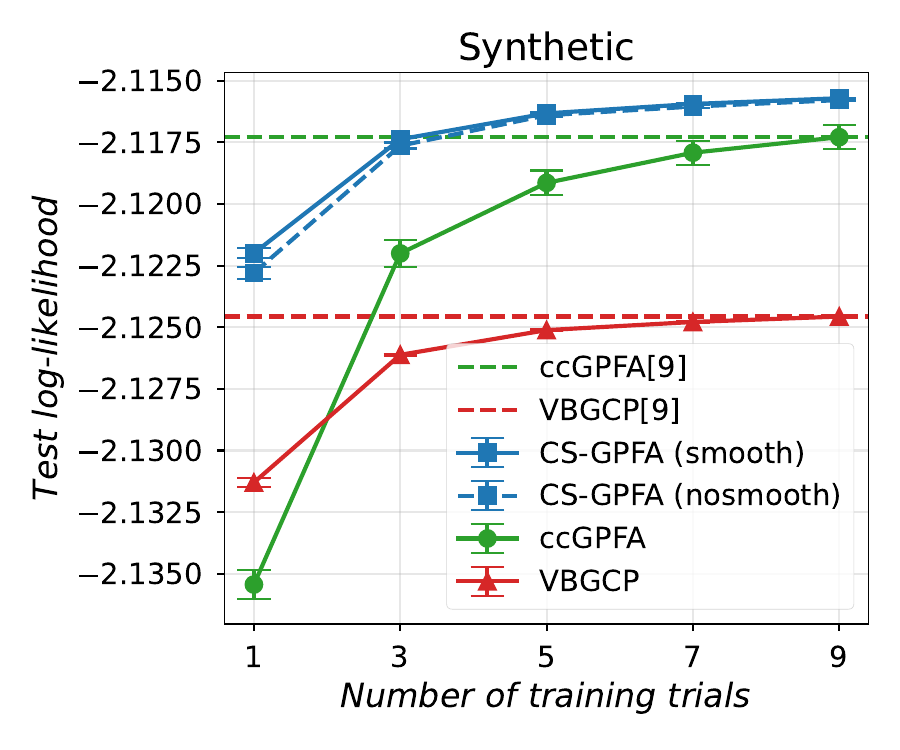}
        \caption{Held-out log-likelihood performance}
    \end{subfigure} \hfill 
    \begin{subfigure}[b]{0.3\textwidth}
        \includegraphics[width=\textwidth]{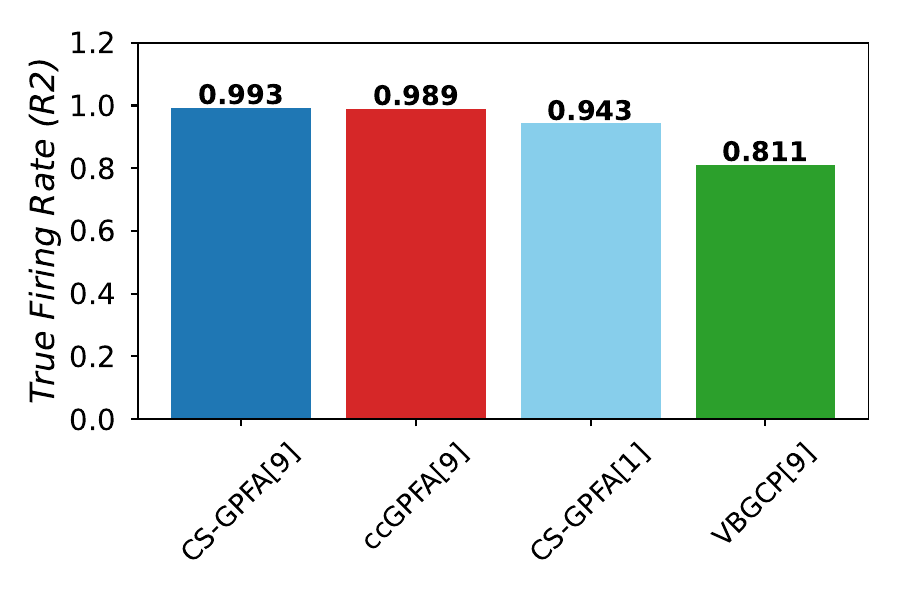}
        \caption{Reconstruction of the true firing rate measure in R2 scores }
    \end{subfigure} \\
    \caption{\textbf{Synthetic Experiment}; error bars represent standard errors computed using 10 independent runs;  \textit{Method[X]} indicates method trained with X held-in trials)}
    \label{appendix:fig:result:synthetic}
\end{figure}

\textbf{Ground truth latent dimensionality} Figure~\ref{appendix:fig:result:synthetic} (a) presents the results for varying choices of latent dimensionality w.r.t the test log-likelihood metric using held-out trials. We can see, for 1-3 latent dimensions, the model shows a steady increase, and for 3 or more, the performance saturates. Despite allocating additional latent dimensions beyond the $D=3$ latent dimensions in the ground truth, the performance does not degrade with increasing dimensions. This behavior reflects the effectiveness of the automatic relevance determination (ARD) prior, which suppresses unnecessary latent dimensions and enables robust model selection.

\textbf{Learning dispersion parameter} Figure~\ref{appendix:fig:result:synthetic} (b) shows \csgpfa's inferred dispersion parameters plotted against the ground truth values. With accurate inference of these parameters, \csgpfa effectively separates ``signal" and ``noise" in the spike data. 

\textbf{Comparison with baselines} As shown in Figure~\ref{appendix:fig:result:synthetic} (c) \csgpfa shows superior performance over the baselines in terms of held-out log likelihood performance, particularly in the single trial setting. With only 3 held-in trials, it shows on par performance compared to the second-place ccGPFA using all the held-in trials. This observation is further supported by the firing rate reconstruction accuracy presented in Figure~\ref{appendix:fig:result:synthetic}(d) which demonstrates that comparable performance can be achieved using only one-tenth of the data (\csgpfa[1]). In contrast, VBGCP showed, in Figure~\ref{appendix:fig:result:synthetic}(c), an increasing trend but relatively poor performance, which could be attributed to the independence assumption it makes along both time and condition axes.

\newpage
\subsection{Visual coding experiment with $N=30$ neurons}
\label{subsec:appendix:visual_coding:fewer_neurons}
Next, we extend the experiment on the vision dataset from Allen Brain Observatory. In this experiment, we uniformly sampled 30 neurons to explore the effectiveness of our method with fewer neurons. Similar to the main experiment, here we run the experiment 10 times with random seeds to select the held-in and held-out trials. 

\textbf{Evaluation Metric} In addition to mean log-likelihood to evaluate test performance, we also report the accuracy of the inferred firing rates compared to the true empirical firing rates in the held-out trials. 

\subsubsection{Results}
\begin{figure}
    \begin{subfigure}[b]{0.4\textwidth}
        \includegraphics[width=\textwidth]{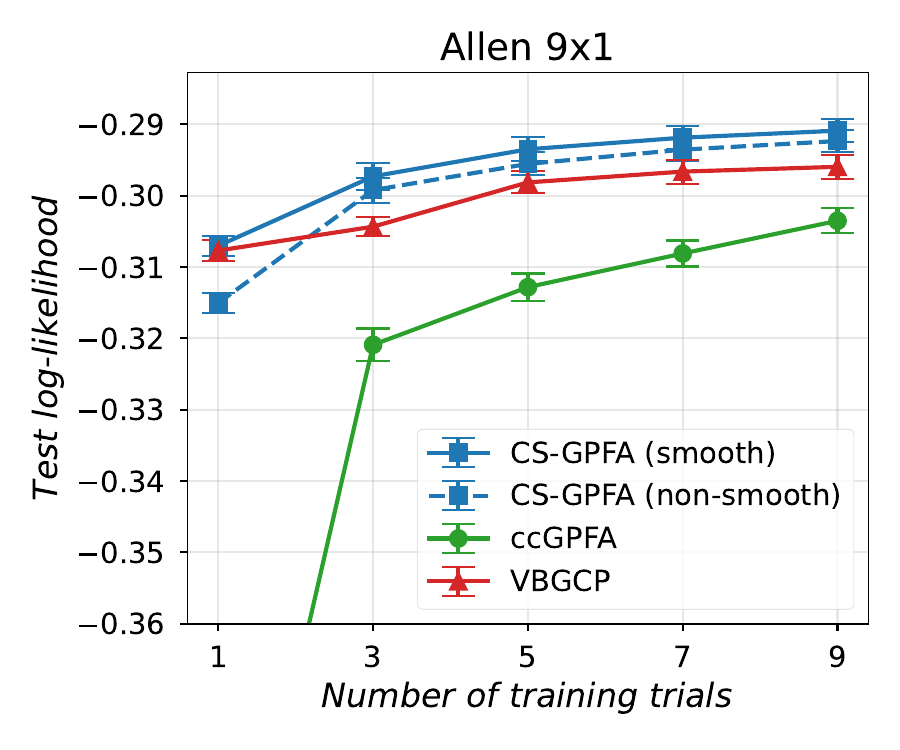}
        \caption{Held-out log-likelihood performance}
    \end{subfigure}\hfill
    \begin{subfigure}[b]{0.4\textwidth}
        \includegraphics[width=\textwidth]{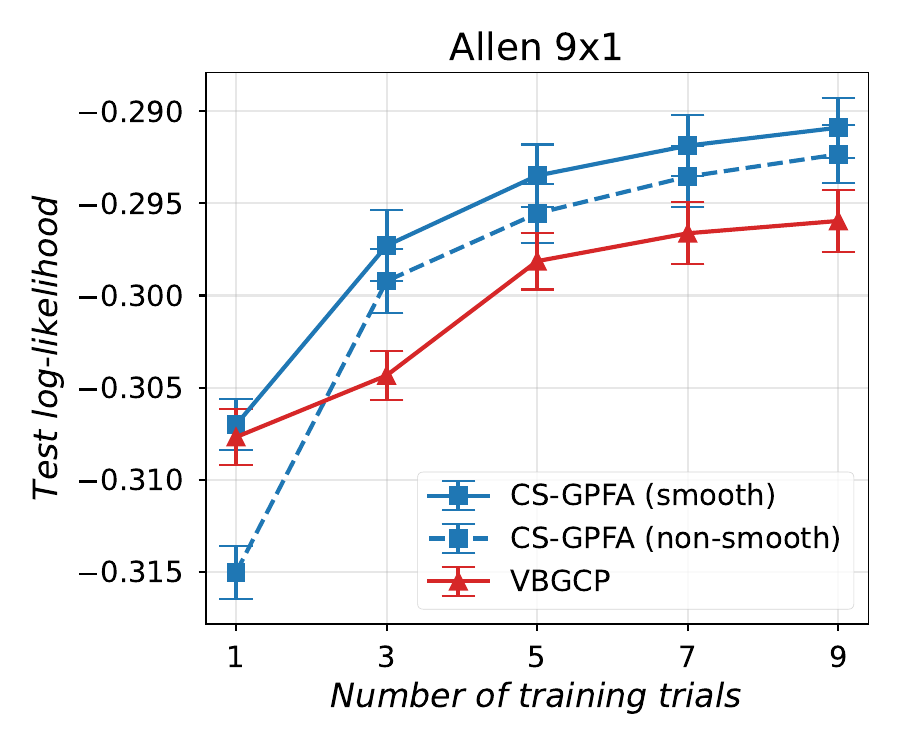}
        \caption{Held-out log-likelihood performance (without ccGPFA for clarity)}
    \end{subfigure} 
    \begin{subfigure}[b]{0.4\textwidth}
        \includegraphics[width=\textwidth]{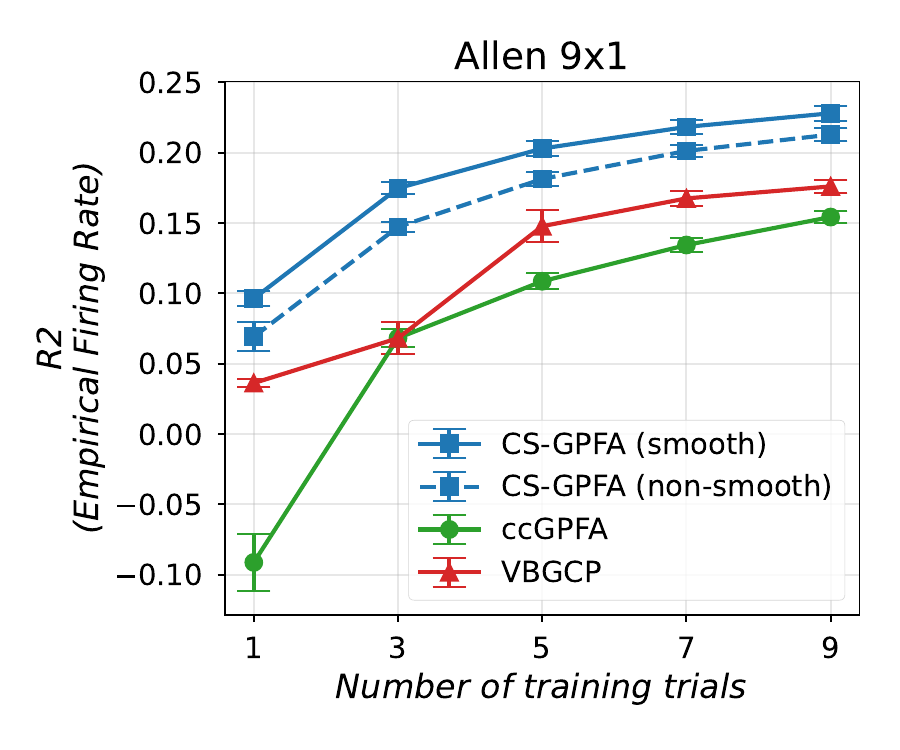}
        \caption{Inferred firing rates comparison with empirical firing rates (R2 scores)}
    \end{subfigure}
    \begin{subfigure}[b]{0.45\textwidth}
      \includegraphics[width=\textwidth]{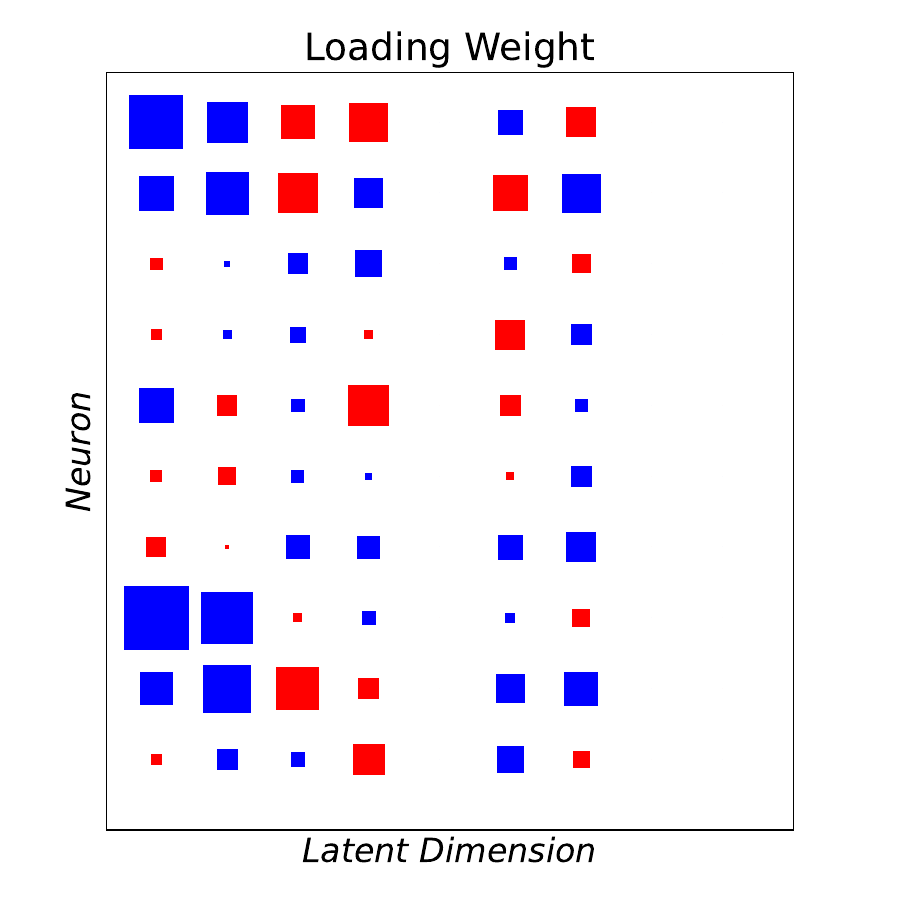}
      \caption{Inferred loading weights}
    \end{subfigure}%
    \hfill
    \begin{subfigure}[b]{0.45\textwidth}
      \includegraphics[width=\textwidth]{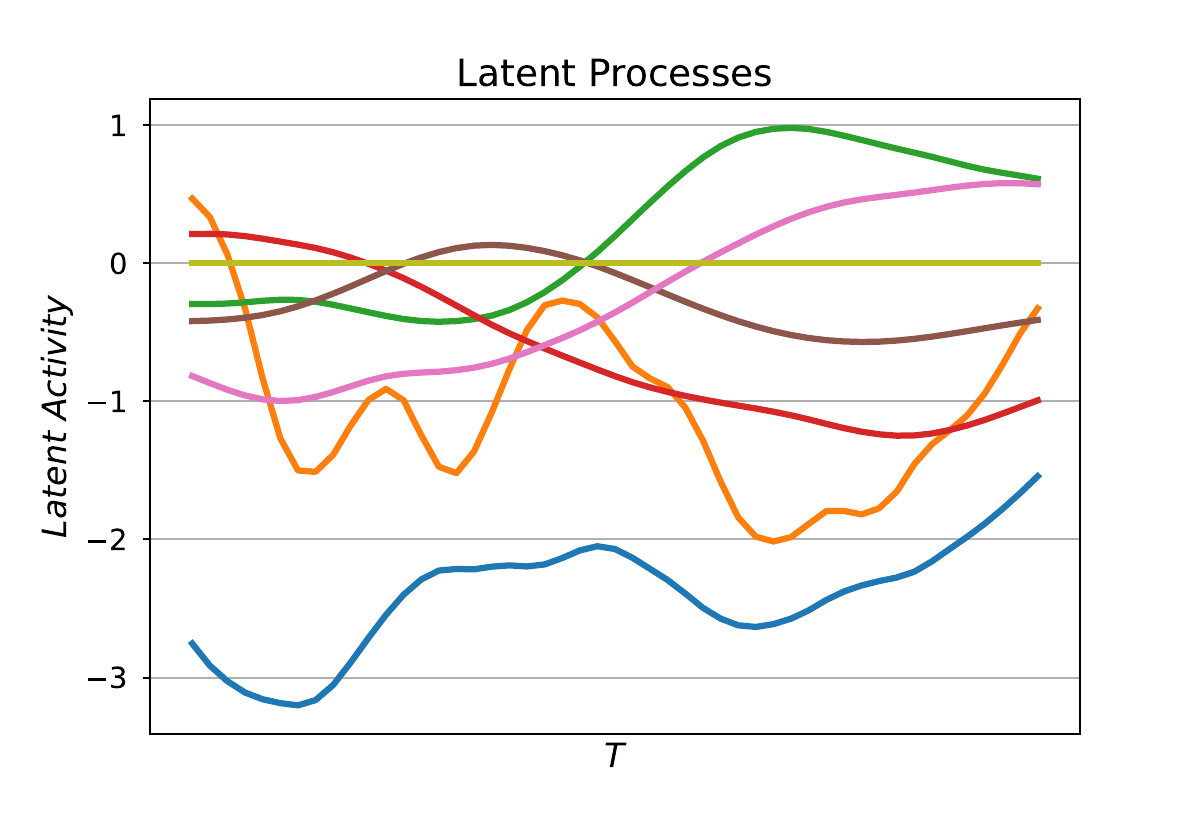}
      \caption{Inferred $D=10$ smooth latent processes (for the first condition)}
    \end{subfigure} \hfill
    \begin{subfigure}[b]{0.45\textwidth}
      \includegraphics[width=\textwidth]{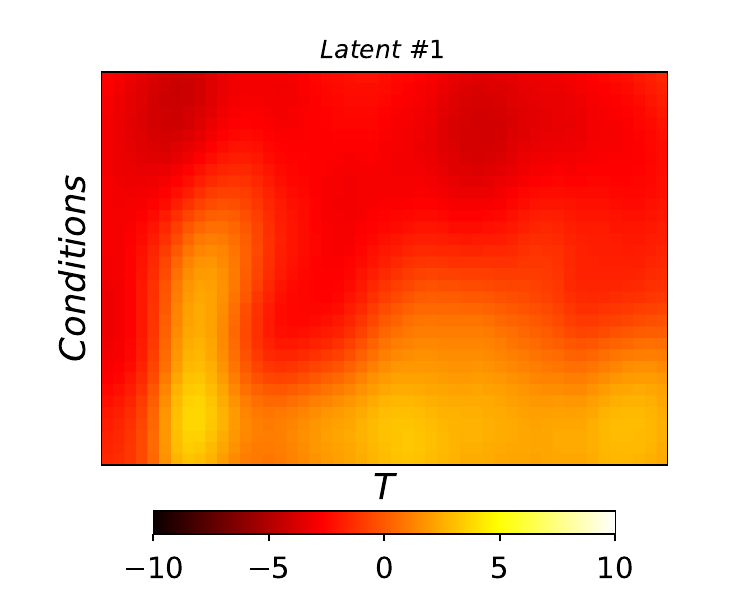}
      \caption{Orthonormalized latent activity of the first principal latent dimension across all conditions }
    \end{subfigure}
    \caption{\textbf{(Extended) Visual Coding Experiment} with $N=30$ Neurons; error bars represent standard errors computed using 10 independent runs. In (f) conditions (y-axis) are arranged from lowest contrast (top) to highest contrast (bottom)} \label{appendix:fig:result:visual_coding_N_30}
\end{figure}

In Figure~\ref{appendix:fig:result:visual_coding_N_30} (a-c), we can see in small trial settings $\{ 1, 3, 5\}$, \csgpfa shows significant improvement w.r.t. both log-likelihood and firing rate metrics. \csgpfa particularly shows superior performance in the single-trial limit compared to ccGPFA, suggesting the utility of smoothness across conditions in improving performance for low-neuron settings. 
Although without the smoothness along the condition axis \nscsgpfa shows an improvement over VBGCP baseline. We attribute this to the GPFA's assumption of smoothness along the temporal axis help the model learn a more regularized latent function.

\textbf{Automatic Relevance Determination} In Figure~\ref{appendix:fig:result:visual_coding_N_30}(d \& e), we can observe \csgpfa retained a total of 6 latent processes out of a prespecified 10 latent processes (including the base activity). In Figure~\ref{appendix:fig:result:visual_coding_N_30} (d), the Hinton diagram shows the loading weights of the first 10 neurons (rows) along the latent dimensions (columns) with sizes of cells representing the magnitude of the weights and colors the sign (red for negative, blue for positive). Here we can see only 6/10 columns (latent dimensions) were retained. %
This showcases how the automatic relevance determination (ARD) prior yields a succinct latent representation. Complementary to the loading weights, Figure~\ref{appendix:fig:result:visual_coding_N_30}(e) shows the inferred latent processes of the first condition, with extraneous latent processes effectively shrunk to 0. And we can see all the non-zero latent processes have corresponding non-zero columns in the Figure~\ref{appendix:fig:result:visual_coding_N_30}(d).      

\textbf{Visualizing the dominant dynamics via orthonormalization} %
Figure~\ref{appendix:fig:result:visual_coding_N_30}(f) shows the first principal latent trajectories across stimulus conditions (rows) over the duration of the trial. As in the main section, we generated these trajectories using \csgpfa{}’s predictions under uniformly sampled contrast levels. Despite using far fewer recorded neurons, the inferred latent states exhibit the same dominant patterns described in the main paper (Figure~\ref{fig:expts:allen_9x1}), including stronger and earlier deviations from baseline activity at higher contrast levels (bottom rows).

\newpage

\subsection{MC Maze experiment}
\label{subsec:appendix:mc_maze_experiment}
\begin{figure}
    \centering
    \begin{subfigure}[b]{0.4\linewidth}
        \includegraphics[width=1.\textwidth]{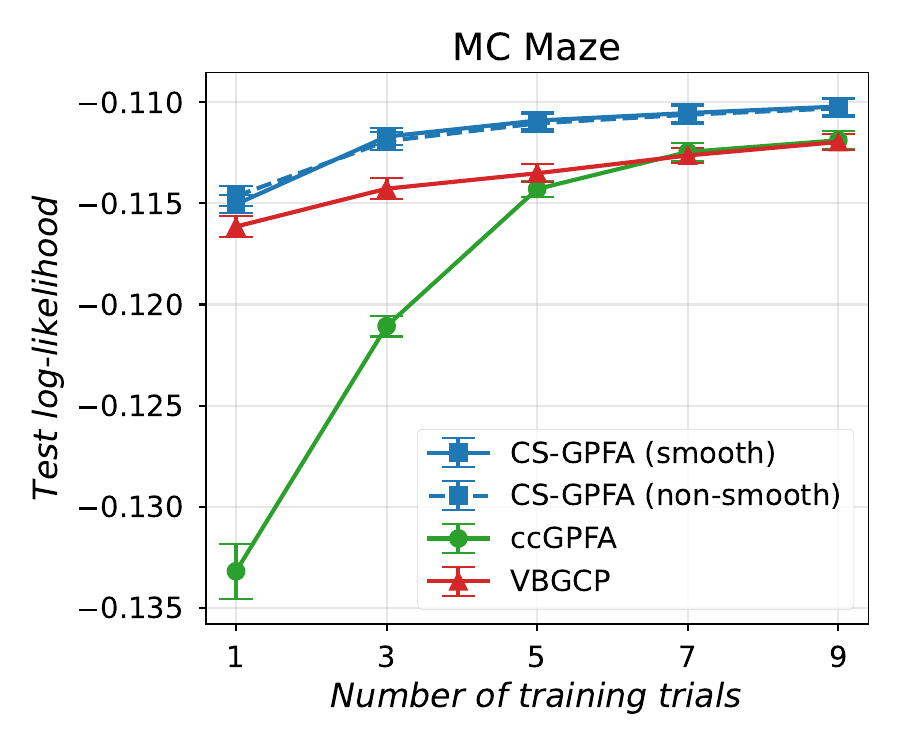}
        \caption{}
    \end{subfigure}\hfill
    \begin{subfigure}[b]{0.4\linewidth}
        \includegraphics[width=1.\textwidth]{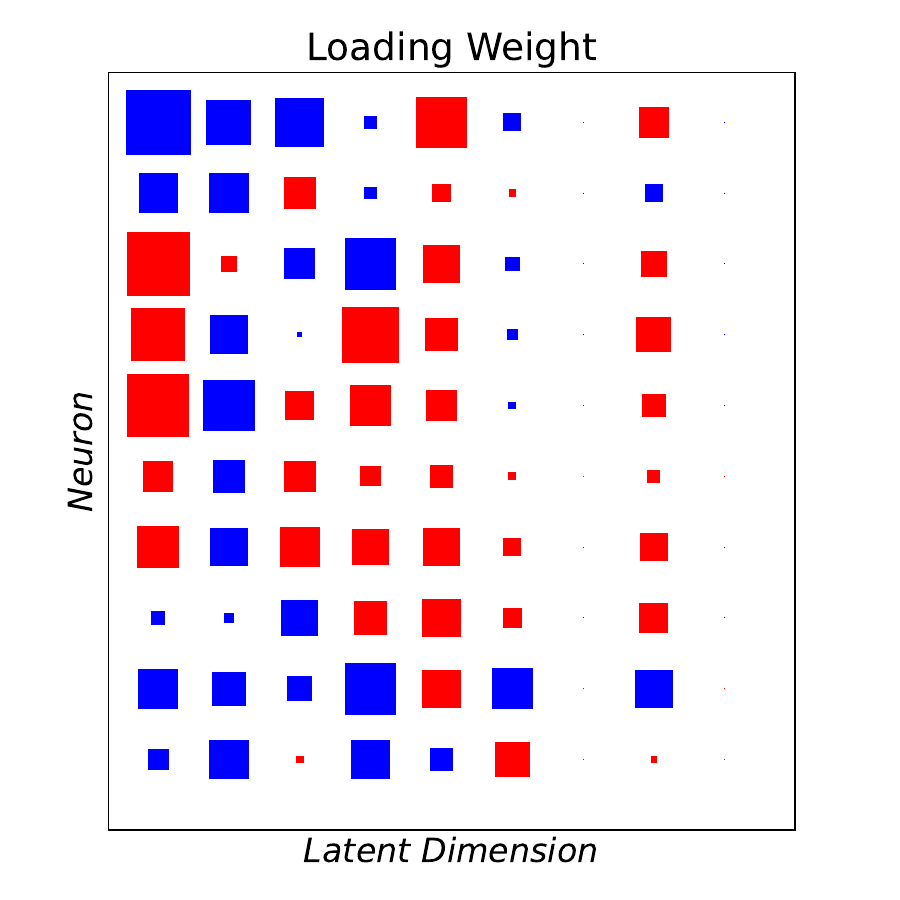}
        \caption{}
    \end{subfigure} 
    \caption{\textbf{MC\_Maze Experiment: 9 reaching angles} 
    \textbf{(a)} Comparison with baselines using mean test log-likelihood performance across varying numbers of training trials. Both \csgpfa methods outperform the baselines by learning accurate representations.
   \textbf{(c)} Inferred loading weights for the first 10 neurons (rows) along each latent dimension (columns); \csgpfa's ARD eliminates 2/10 superfluous latent to yield a concise representation.
    }
    \label{appendix:fig:result:mc_maze}
\end{figure}

Following the experiment setup presented in Section~\ref{subsection:experiment:visual_coding_experiment}, we run the experiment on a new center-out reaching task dataset. The results of the experiments are presented in Figure~\ref{appendix:fig:result:mc_maze}.

\textbf{Dataset} The MC\_Maze dataset \citep{churchland2010cortical, pei2021neural} consists of neural recordings from the primary motor and dorsal premotor cortices of a monkey performing a center-out reaching task, in which the subject reaches toward a visual target while navigating around a virtual maze. The task is presented under different configurations corresponding to distinct reach orientations or movement angles in the range $[-\pi,\pi]$.

For our analysis, we extracted 500 ms segments of neural activity, spanning 130 ms before to 370 ms after movement onset. The resulting dataset includes recordings from 9 reach-angle configurations, each repeated for 18 trials. Spike trains were discretized using a 10 ms bin width.

\textbf{Results} Figure~\ref{subsec:appendix:mc_maze_experiment}(a) compares \csgpfa with the baseline methods. Overall, both the smooth and non-smooth variants of \csgpfa{} outperform the baselines, demonstrating the benefit of coupling latent representations across time and conditions when learning accurate latent factors. Although \csgpfa{} offers comparable or slightly improved performance relative to \nscsgpfa, the gains are less pronounced than in other experiments. This is likely because the conditions (reaching angles) in this dataset are relatively sparse, requiring more samples to reliably capture the relationships among configurations.

\newpage
\subsection{Active learning experiment (contd)}

\begin{figure}

    \centering
    \begin{subfigure}[b]{.6\textwidth}
    {
        \includegraphics[width=1.\linewidth]{results_allen_9x4_active_learning_entropy_us.pdf}
        \\
        \includegraphics[width=1.\linewidth]{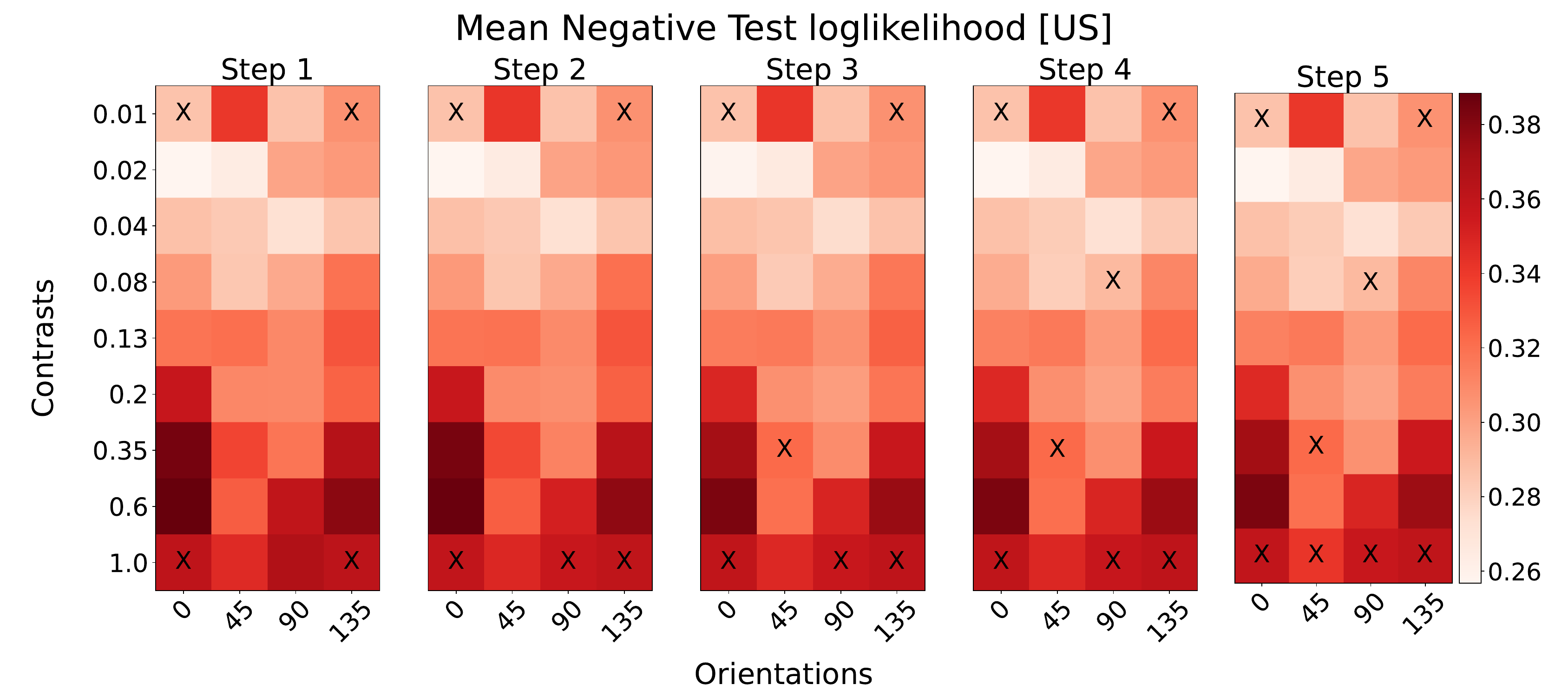}
        
        \subcaption{\textbf{(Uncertainty Sampling)} Heatmap plots for predictive entropy (\textbf{top}) and negative held-out test log-likelihood (\textbf{bottom}) at each step of the algorithm. 
        }
    }
    \end{subfigure}
    \begin{subfigure}[b]{0.6\textwidth}
    {
        \includegraphics[width=1.\linewidth]{results_allen_9x4_active_learning_entropy_reductions_ig.pdf}
        \\
        \includegraphics[width=1.\linewidth]{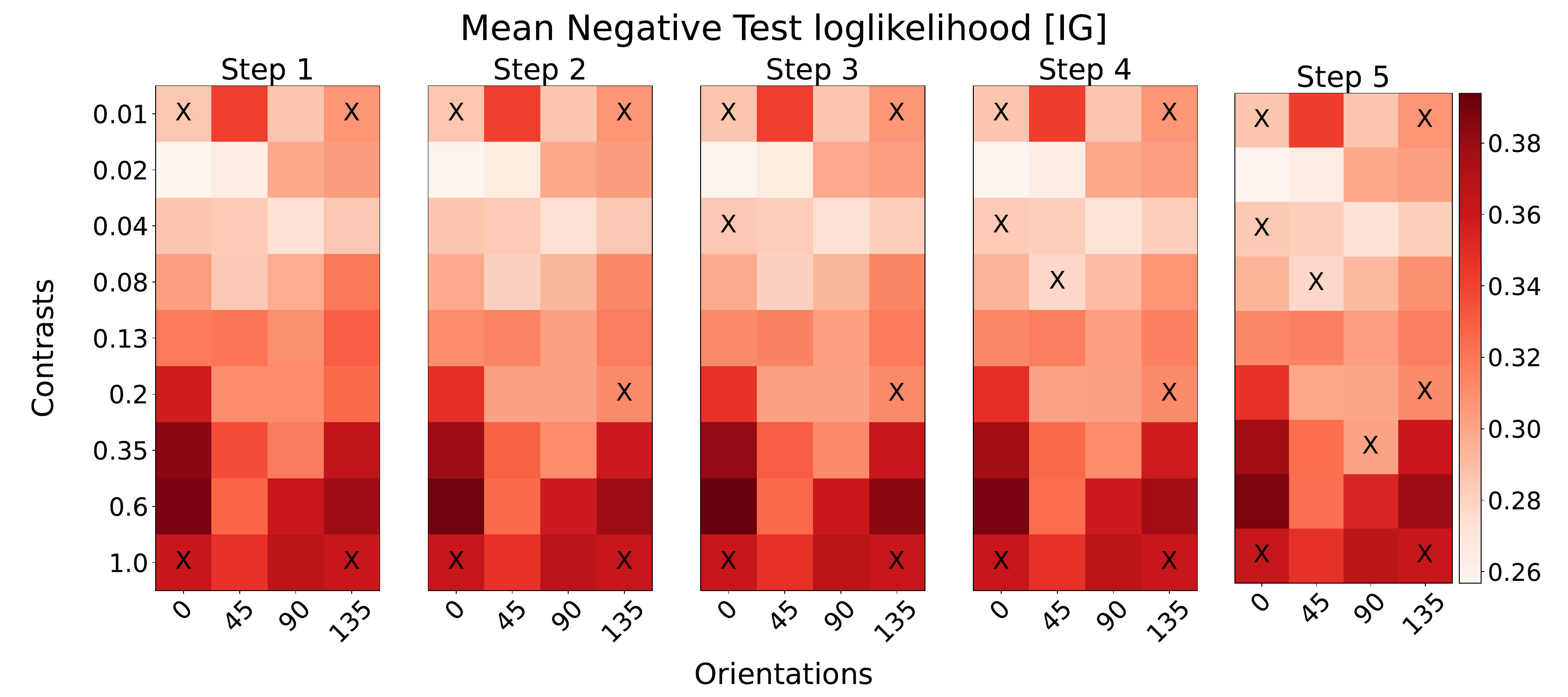}

        \subcaption{\textbf{ (Information Gain)} Heatmap plots for entropy reduction (information gain) (\textbf{top}) and negative held-out test log-likelihood (\textbf{bottom}) at each step of the algorithm. }
    }
    \end{subfigure}
    \caption{\textbf{Active Learning Experiment}}

    \label{fig:result:active-learning-2}
\end{figure}

\newpage
\subsection{Resources}

We conducted the experiments on a  desktop machine with a 
AMD EPYC 9655 processor and 16GB of memory. 
We used Python (v3.11), Jax (v0.4.14) and the BayesNewton library\footnote{https://github.com/AaltoML/BayesNewton} \citep{wilkinson2023bayes}.